\documentclass[journal]{IEEEtran}

\usepackage{graphicx}
\usepackage[cmex10]{amsmath}
\usepackage{mathptmx} 
\usepackage{subfigure}
\usepackage{epstopdf}
\usepackage[left=1in, right=0.7in, bottom=0.4in, top=1in]{geometry}
\usepackage[english]{babel}
\usepackage{dcolumn}
\usepackage{multirow}

\usepackage{color}

\usepackage[]{cite}

\newcommand{\argmin}[1]{\underset{#1}{\operatorname{arg}\,\operatorname{min}}\;}

\begin{document}

\title{A Comparative Study of Reservoir Computing for Temporal Signal Processing}

\author{Alireza~Goudarzi$^1$,
        Peter~Banda$^2$,~Matthew~R.~Lakin$^1$,~Christof Teuscher$^3$,
        and~Darko~Stefanovic$^1$
\thanks{$^1$ Department
of Computer Science, University of New Mexico, Albuquerque,
NM 87131, USA e-mail: alirezag@cs.unm.edu.}
\thanks{$^2$ Department of Computer Science, Portland State University, Portland, OR 97207, USA.}
\thanks{$^3$ Department of Electrical and Computer Engineering, Portland State University, Portland, OR 97207, USA.}}

\maketitle

\begin{abstract}
 
Reservoir computing (RC) is a novel approach to 
time series prediction using recurrent neural networks. In RC, an input signal perturbs the intrinsic dynamics of a 
medium called a reservoir. A readout layer is then trained to reconstruct a target 
output from the reservoir's state. The multitude of RC architectures and evaluation 
metrics poses a challenge to both practitioners and theorists who study the task-solving performance and computational power of RC. In addition, in contrast to 
traditional computation models, the reservoir is a dynamical system in which 
computation and memory are inseparable, and therefore hard to analyze. Here, we compare 
echo state networks (ESN), a popular RC architecture, with tapped-delay lines (DL) and  
nonlinear autoregressive exogenous (NARX) networks, which we use to model systems with
limited computation and limited memory respectively.
We compare the performance of the three systems while computing three common benchmark time 
series: H{\'e}non Map, NARMA10, and NARMA20. We find that the role of the 
reservoir in the reservoir computing paradigm goes beyond providing a memory of the 
past inputs. The DL and the NARX network have higher memorization capability, but fall short of the generalization power of the ESN.
\end{abstract}


%
\IEEEpeerreviewmaketitle

\section{Introduction}
%
%
%
%

Reservoir computing  is a recent development in recurrent 
neural 
network research with applications to temporal pattern recognition~\protect
\cite{Schrauwen07anoverview}. RC's performance in time series processing
tasks and its flexible implementation has made it an intriguing concept in machine 
learning and unconventional computing 
communities~
\cite{1556091,Wyffels20101958,Jaeger:
2003p1447,Lukosevicius:2009p1443,0957-4484-24-38-384004,866618,0957-4484-24-38-384010,Obst2013,Paquot:
2012fk,Fernando2003,Goudarzi2013a}. In this paper, we functionally compare the performance of reservoir 
computing with linear and nonlinear autoregressive methods for temporal signal 
processing to develop a baseline for understanding memory and information processing in reservoir computing.

\begin{figure*}
\centering
\includegraphics[]{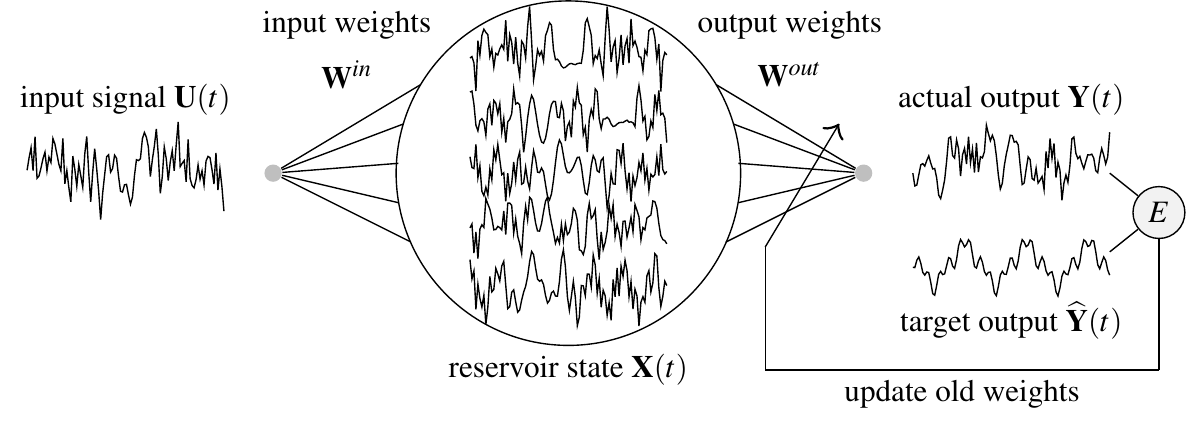}
\caption{Computation in a reservoir computer. The reservoir is made up of a dynamical neural network with randomly assigned weights. The state of the nodes are represented by ${\bf X}(t)$. The input signal ${\bf U}(t)$ is fed 
into every reservoir node $i$ with a corresponding weight $w^{in}_i$ 
denoted with weight column vector ${\bf W}^{in}=[w^{in}_i]$. Reservoir nodes are 
themselves coupled with each other using the weight matrix ${\bf W}^{res}
=[w^{res}_{ij}]$, where $w^{res}_{ij}$ is the weight of the connection from node $j$ to node $i$.}
\label{fig:fig_0}
\end{figure*}

In reservoir computing, a high-dimensional dynamical core called a {\em reservoir} is 
perturbed with an external input. The reservoir states are then linearly combined to 
create the output. The readout parameters can be calculated by performing regression 
on the state of a teacher-driven reservoir and the expected teacher output. Figure~\ref{fig:fig_0} shows a sample RC architecture. Unlike other forms 
of neural computation, computation in RC 
takes place within the transient dynamics of the reservoir.
The computational power of the reservoir is attributed to a short-term memory 
created by the reservoir~\protect\cite{Jaeger02042004} and the ability to preserve the 
temporal information
from distinct signals over time~\cite{Maass:2002p1444,Natschlaeger2003}. Several studies attributed this property to the dynamical regime of the reservoir and showed it to be optimal when the system operates in the critical dynamical regime---a regime 
in which perturbations to the system's trajectory in its phase space neither spread nor 
die out~\protect\cite{Natschlaeger2003,Bertschinger:2004p1450,snyder2013a,
4905041020100501,Boedecker2009}. The reason for this observation remains unknown. 
 Maass et al.~\protect\cite{Maass:2002p1444} proved that given the two 
properties of {\em separation} and {\em approximation}, a reservoir system is capable 
of approximating any time series. The separation property ensures that the reservoir 
perturbations from distinct signals remain distinguishable whereas the approximation 
property ensures that the output layer can approximate any function of the reservoir 
states to an arbitrary degree of accuracy. Jaeger~\protect\cite{Jaeger:2002p1445} proposed that an ideal reservoir needs to have the so-called echo state property (ESP), which means that the reservoir 
states asymptotically depend on the input and not the initial state of the reservoir. It 
has also been suggested that the reservoir dynamics acts like a spatiotemporal 
kernel, projecting the input signal onto a high-dimensional feature space~\protect
\cite{Hermans:2011fk,Lukosevicius:2009p1443}. However, unlike in kernel methods, 
the reservoir explicitly computes the feature space.

 RC's robustness to the underlying implementation as well as its 
efficient training algorithm makes it a suitable choice for time series analysis 
\cite{springerlink:10.1007}. However, despite more than a decade of research in RC 
and many success stories, its wide-spread adoption is still forthcoming for three main 
reasons: first, the lack of theoretical understanding of RC's working and its 
computational power, and 
second,  the absence of a unifying implementation framework and performance 
analysis results, and thirdly, missing  comparison with conventional methods.

\section{Objectives}
Our main objective is to compare time series computing in the RC paradigm, in which 
memory and computation are integrated, with two basic time series computing methods:
first, a device with perfect memory and no 
computational power, ordinary linear regression on tapped-delay line (DL); and second, a device with limited memory and arbitrary computational 
power, a nonlinear autoregressive exogenous (NARX) neural network. This is a first step toward
a systematic investigation of topology, memory, computation, and 
dynamics in RC. In this article we restrict ourselves to ESNs with a fully connected 
reservoir and input layer. We study the performance of ESN and autoregressive model on solving 
three time series problems: computing the 10th order NARMA time series~\protect
\cite{846741}, the 20th order NARMA time series~\protect\cite{5629375}, and the 
H{\'e}non Map~\protect\cite{Henon:1976fk}. We also provide  performance results using 
several variations of the  mean squared error (MSE) and symmetric mean absolute 
percentage (SAMP) error  to make our results accessible to the broader neural 
network and time series analysis research communities. Our systematic comparison 
between the ESN and autoregressive model provides solid evidence that the reservoir 
in the ESN performs non-trivial computation and is not just a memory device.

\section{A Brief Survey of Previous Work}
 The first conception of the RC paradigm in the recurrent neural network (RNN) 
community was Jaeger's  echo state network (ESN)~\protect\cite{Jaeger:
2001p1442}. In this early ESN, the reservoir consisted of $N$  fully interconnected sigmoidal 
nodes. The reservoir connectivity was represented by a weight matrix with  elements
sampled from a uniform distribution in the interval $[-1,1]$. The weight 
matrix was then rescaled to have a spectral radius of $\lambda<1$, a sufficient condition for ESP. 

The input signal was connected to all the reservoir nodes and their weights were  
randomly assigned from the set $\{-1,1\}$. Later, Jaeger~\protect\cite{Jaeger:2001p1446,Jaeger:
2002p1445} proposed that the sparsity of the connection weight matrix would improve 
performance and therefore only $20\%$ of the connections were assigned  weights 
from the set $\{-47,47\}$.Verstraeten et al.~\protect\cite{verstraeten2007} used a  
$50\%$ sparse reservoir and a normal distribution for the  connection weights, 
and scaled the weight matrix posteriori to ensure the ESP; also, only $10\%$ of the 
nodes were connected to the input. This study indicated that, contrary to the earlier report 
by Jaeger~\cite{Jaeger:2001p1442}, the performance of the reservoir was sensitive to 
the spectral radius and showed optimality for $\lambda\approx 1.1$. Venayagamoorthy 
and Shishir~\protect\cite{Venayagamoorthy2009861} demonstrated experimentally 
that the spectral radius also affects training time, but, they did not study spectral 
radii larger than one. Gallicchio and Micheli~\protect\cite{Gallicchio2011440} provided 
evidence that the sparsity of the reservoir has a negligible effect on ESN 
performance, but depending on the  task, input weight heterogeneity can 
significantly improve performance. B\"using et al.~\protect
\cite{4905041020100501} reported, from private communication with Jaeger, that 
different reservoir structures, such as the scale-free and the small-world topologies, 
do not have any significant effect on ESN performance. Song and Feng~\protect
\cite{Song20102177} demonstrated that in ESNs, with complex network reservoirs, high average path length and low clustering 
coefficient improved the performance. This finding is at odds with what has been observed in complex 
cortical circuits~\protect\cite{Bullmore:2009vn} and other studies of ESN~\protect
\cite{10.3389/conf.fncom.2011.53.00177}. 

Rodan and Tino~\protect\cite{5629375} 
studied an ESN model with a very simple reservoir consisting of nodes that are 
interconnected in a cycle with homogeneous input weights and homogeneous 
reservoir weights, and showed that its performance can be made arbitrarily close to 
that of the classical ESN. This finding addressed for the first time  concerns about 
the practical use of  ESNs in embedded systems due to their complexity 
\cite{1556091}. Massar and Massar \protect\cite{PhysRevE.87.042809} formulated a 
mean-field approximation to the ESN reservoir and demonstrated that the optimal 
standard deviation of a normally distributed weight matrix $\sigma_w$ is an inverse 
power-law of the reservoir size $N$ with exponent $-0.5$. However, this optimality is 
based on having critical dynamics and not task-solving performance.

\begin{figure*}
\centering
\subfigure[]{
\includegraphics[width=2.0in]{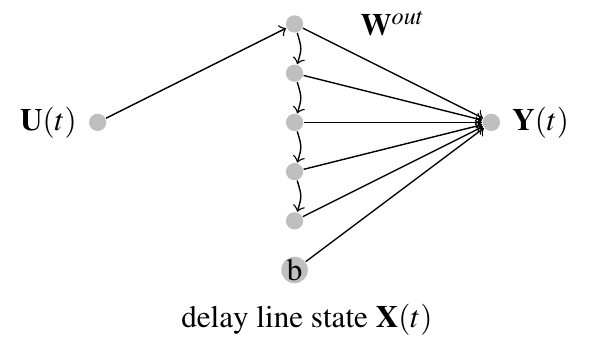}
\label{fig:figdlarch}
}
\subfigure[]{
\includegraphics[width=2.0in]{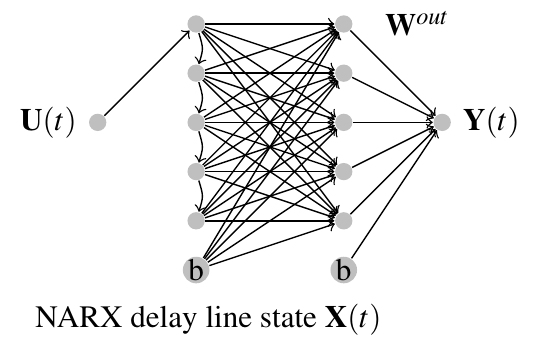}
\label{fig:fignnarch}
}
\subfigure[]{
\includegraphics[width=2.0in]{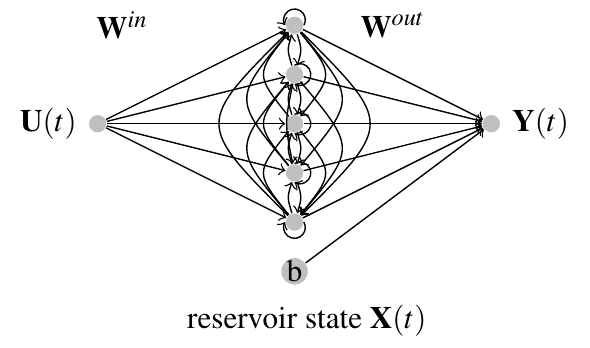}
\label{fig:figrcarch}
}
\caption{Architecture of delay line with a linear readout, the NARX neural network, and the ESN.}
\label{fig:Fig_arch}
\end{figure*}

\section{Models}

To understand reservoir computation, we compare its 
behavior with a system with perfect memory and no computational power and a 
system with limited memory and arbitrary computational power.

We choose  delay 
line systems,  NARX neural networks, and  echo state networks as described below. 
We use $U(t)$, $Y(t)$, and $\widehat{Y}(t)$ to denote the time-dependent input 
signals, the time dependent output signal, and the time-dependent target signal, 
respectively. 

\subsection{Delay line}
\label{sec:dl}
A tapped-delay line (DL) is a simple system that allows us to access a delayed version 
of a signal.  To 
compare the computation in a reservoir with the DL, we use a linear readout layer and 
connect it to read the states of the DL. Figure~\ref{fig:figdlarch} is a schematic for this 
architecture. Note that this architecture is different from the delay line used in~\protect
\cite{5629375} in that the input is only connected to a single unit in the delay line.
The delay units do not perform any computation. The system is then fed with a teacher 
input and the weights are trained using an ordinary linear regression on the teacher 
output as follows:

\begin{equation}
{\bf W}^{out}=({\bf X}^T\cdot {\bf X})^{-1}\cdot {\bf X}^T\cdot \widehat{\bf Y},
\label{eq:regeq}
\end{equation}
where each row of ${\bf X}$ represents the state of the DL at a specific time 
$X(t_0)$ and the rows of $\widehat{\bf Y}$ are the teacher output at for the 
corresponding time $Y(t_0)$. Note that the DL states are augmented with a bias 
unit with a constant value $b=1$. Initially, all the DL states are set to zero.
\subsection{NARX Network}
\label{sec:narx}
The NARX network is an autoregressive neural network architecture with a tapped 
delay input and one or more hidden layers. Both hidden layers and the output layer 
are provided with a bias input with constant value $b=1$. We use $\tanh$ as the 
transfer function for the hidden layer and a linear transfer function for the output layer.  The 
network is trained using the Marquardt algorithm~\protect\cite{329697}.  This architecture 
performs a nonlinear regression on the teacher output using the previous values of the 
input accessible through the tapped delay line. A schematic of this architecture is 
given in Figure~\ref{fig:fignnarch}. Since we would like to study the effect of regression 
complexity on the performance, we fix the length of the tapped 
delay to 10 and vary the number of hidden nodes. 
\subsection{Echo State Network}
\label{sec:rc}
In our ESN, the reservoir consists of a fully connected network of $N$ nodes extended 
with a constant bias node $b=1$. The input and the output nodes are connected to all 
the reservoir nodes. The input weight matrix is an $I\times N$ matrix ${\bf W}^{in}
=[w^{in}_{i,j}]$, where $I$ is the number of input 
nodes and $w^{in}_{j,i}$ is the 
weight of the connection from input node $i$ to reservoir node $j$. 
The connection weights inside the reservoir are represented by an
 $N\times N$ matrix ${\bf W}^{res}=[w^{res}_{j,k}]$, where $w^{res}_{j,k}$ is the 
weight from node $k$ to node $j$ in the reservoir. The output weight matrix is 
an $(N+1(\times O$ matrix ${\bf W}^{out}=[w^{out}_{l,k}]$, where $O$ is the number 
of output nodes and $w^{out}_{l,k}$ is the weight of the connection from 
reservoir node $k$ to output node $l$. All the weights are samples of 
i.i.d. random variables from a normal distribution with mean $\mu_w=0$ and 
standard deviation $\sigma_w$.  We represent the 
time-varying input signal by an $I$th order column vector ${\bf U}(t)=[u_i(t)]$, the 
reservoir state by an $N$th order column vector ${\bf X}(t)=[x_j(t)]$, and the 
generated output by an $O$th order column vector ${\bf Y}(t)=[y_l(t)]$. We  compute 
the  time evolution of each reservoir node in discrete time 
as:
\begin{equation}
x_j(t+1)=\tanh({\bf W}^{res}_j\cdot {\bf X}(t)+{\bf W}^{in}\cdot {\bf U}(t)),
\end{equation}
where $\tanh$ is the nonlinear transfer function of the reservoir nodes and ${\bf W}^{res}_j$ is the $j$th 
row of the reservoir weight matrix. The reservoir output is then given by:
\begin{equation}
{\bf Y}(t)={\bf W}^{out}\cdot{\bf X}(t).
\end{equation}
 We train the output weights to 
minimize the squared output error $E=||{\bf Y}(t)-{\bf \widehat{Y}}(t)||^2$ given the target output 
${\bf \widehat{Y}}(t)$. As with DL, the output weights are calculated using the ordinary 
linear regression given in Equation~\ref{eq:regeq}. 
\section{Evaluation Methodology}
The study of task-solving performance and analysis of computational power in RC is a 
major challenge because there are a 
variety of RC architectures, each with a unique set of parameters that potentially affect 
the performance. The optimal RC parameters are task-dependent and  must be 
adjusted experimentally. Furthermore, not all studies use the same tasks and the same 
performance metrics to evaluate their results. In addition, in contrast to classical 
computation models in which a programmed automaton acts on a storage device, RC is 
a dynamical system in which memory and computing are inseparable parts of  a 
single phenomenon. In other words, in RC the same dynamical process that performs 
computation also retains the memory of the previous results and inputs. Thus, it is not 
clear how much of the RC's performance can be attributed to its memory capacity 
and how much to its computational power. As a way of approaching this issue, we attempt to create a 
functional comparison between the ESN,  DL, and  NARX networks.

We choose 
three temporal tasks for our evaluation: the H{\'e}non Map, the NARMA 10 time series, 
and the NARMA 20 time series. These tasks vary in increasing order in their time lag 
dependencies and the number of terms involved and thus let us compare the 
performance of our systems based on the memory and computational requirements 
for task solving.  We measure the performance using three variations of MSE and a 
SAMP error measure to allow easy comparison with related work.
\subsection{Tasks}
\label{sec:tasks}
\subsubsection{H{\'e}non Map Time Series}
This time series is generated by the following system:
\begin{equation}
y_t=1-1.4y_{t-1}^2+0.3y_{t-2}+z_t,
\end{equation} 
where $z_t$ is a white noise term with standard deviation $0.001$. This is an example of a task that requires limited computation and memory, and can therefore be used as a baseline to evaluate ESN performance.
\subsubsection{NARMA 10 Time Series}
Nonlinear autoregressive moving average (NARMA) is a discrete-time temporal task 
with $10$th-order time lag. To simplify the notation we use $y_t$ to denote $y(t)$. The 
NARMA 10 time series is given by: 
\begin{equation}
y_t=\alpha y_{t-1}+\beta y_{t-1} \sum_{i=1}^{n}y_{t-i}+\gamma u_{t-n}u_{t-1}+\delta,
\end{equation}
where $n=10$, $\alpha=0.3, \beta=0.05, \gamma=1.5, \delta=0.1$. The input $u_t$ is 
drawn from a uniform distribution in the interval $[0,0.5]$. This task presents a challenging 
problem to any computational system because of its nonlinearity and dependence on 
long time lags. Calculating the task is trivial if one has access to a device capable of 
algorithmic programming and perfect memory of both the input and the outputs of up 
to 10 previous time steps. This task is often used to evaluate the memory capacity 
and computational power of ESN and other recurrent neural networks.
\subsubsection{NARMA 20 Time Series} NARMA 20 requires twice the memory 
and computation compared to NARMA 10 with an additional nonlinearity because of the 
saturation function $tanh$. This task is very unstable and the saturation function 
keeps its values bounded. NARMA 20 time series is given by:
\begin{equation}
y_t=\tanh(\alpha y_{t-1}+\beta y_{t-1} \sum_{i=1}^{n}y_{t-i}+\gamma u_{t-n}u_{t-1}+
\delta),
\end{equation}

where $n=20$, and the rest of the constants are set as in NARMA 10.
\subsection{Error Calculation}
A challenge in comparing results across different studies is the way each study 
evaluates its results. In the case of time series analysis, each study may use a 
different error calculation to measure the performance of the presented methods. We 
present three different error calculations commonly used in the time series analysis 
literature. We use $y$ to refer to the time-dependent output and $\widehat{y}$ to refer 
to the target output. The expectation operator $\langle\cdot\rangle$ refers to the time 
average of its operand. 
\subsubsection{Root normalized mean squared error} The most commonly used 
measure is a root normalized mean squared error (RNMSE) calculated as follows:
\begin{equation}
RNMSE = \sqrt{\frac{\langle (y - \widehat{y})^2\rangle}{\sigma_{\widehat{y}}w^2}}.
\end{equation}
 Here $\sigma_{\widehat{y}}^2$ is the standard deviation of the target output over 
time. In some studies this calculation is used without taking the square root, in which 
case it is simply called a normalized mean squared error (NMSE). 
 \subsubsection{Normalized root mean squared error} A variant of the normalized error is 
the  normalized root mean square error (NRMSE), also known as normalized root mean 
squared deviation (NRMSD). It is calculated as follows:
\begin{equation}
NRMSE = \frac{\sqrt{\langle (y - \widehat{y})^2\rangle}}{\text{max}(\widehat{y})-
\text{min}(\widehat{y})}.
\end{equation}
In this variant, the error is normalized by the width of the range covered by the target 
signal. Both RNMSE and NRMSE attempt to normalize the error between $0$ 
and $1$. However, if the distance between the output and the target output is larger 
than the standard deviation of the target output or its range, they may produce an 
error value larger than $1$. 
\subsubsection{Symmetric absolute mean percentage error} The symmetric absolute 
mean percentage (SAMP) error, on the other hand, is guaranteed to produce an error 
value between $0\%$ and $100\%$. SAMP is given by: 
\begin{equation} 
SAMP=100\times \left\langle \frac{|y-\widehat{y}|}{y+\widehat{y}} \right\rangle.
\end{equation}

Throughout the rest of the paper, we use the RNMSE error to produce the plots and make 
our comparison between the three systems. We have tabulated the results using the other 
metrics in the Appendix~\ref{sec:appb}.
 
\begin{figure*}[ht!]
\centering
\subfigure[H{\'e}non Map error surface]{
\includegraphics[width=2.0in]{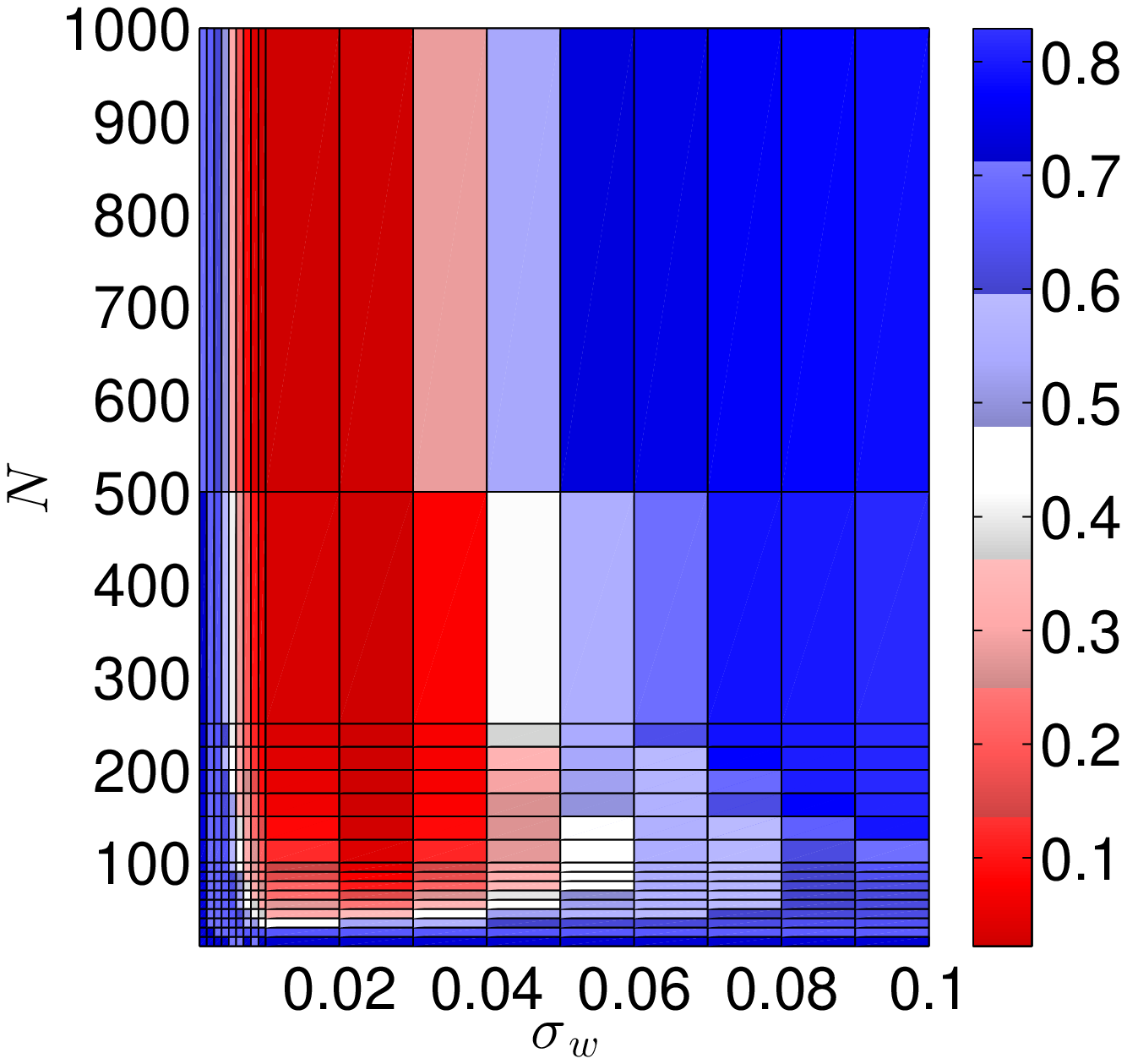}
\label{fig:surfhanon}
}
\subfigure[NARMA 10 error surface]{
\includegraphics[width=2.0in]{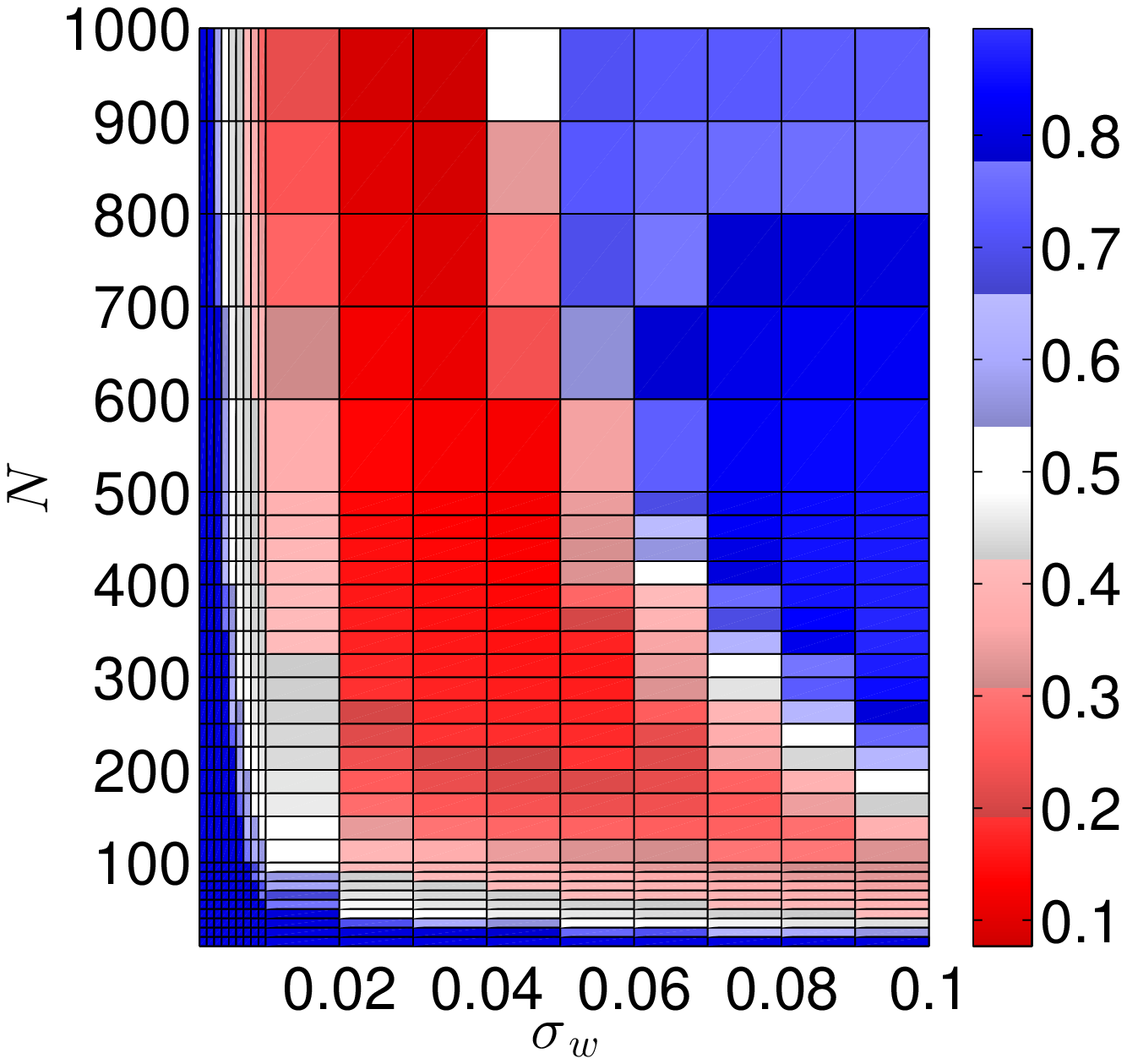}
\label{fig:surfnarma10}
}
\subfigure[NARMA 20 error surface]{
\includegraphics[width=2.0in]{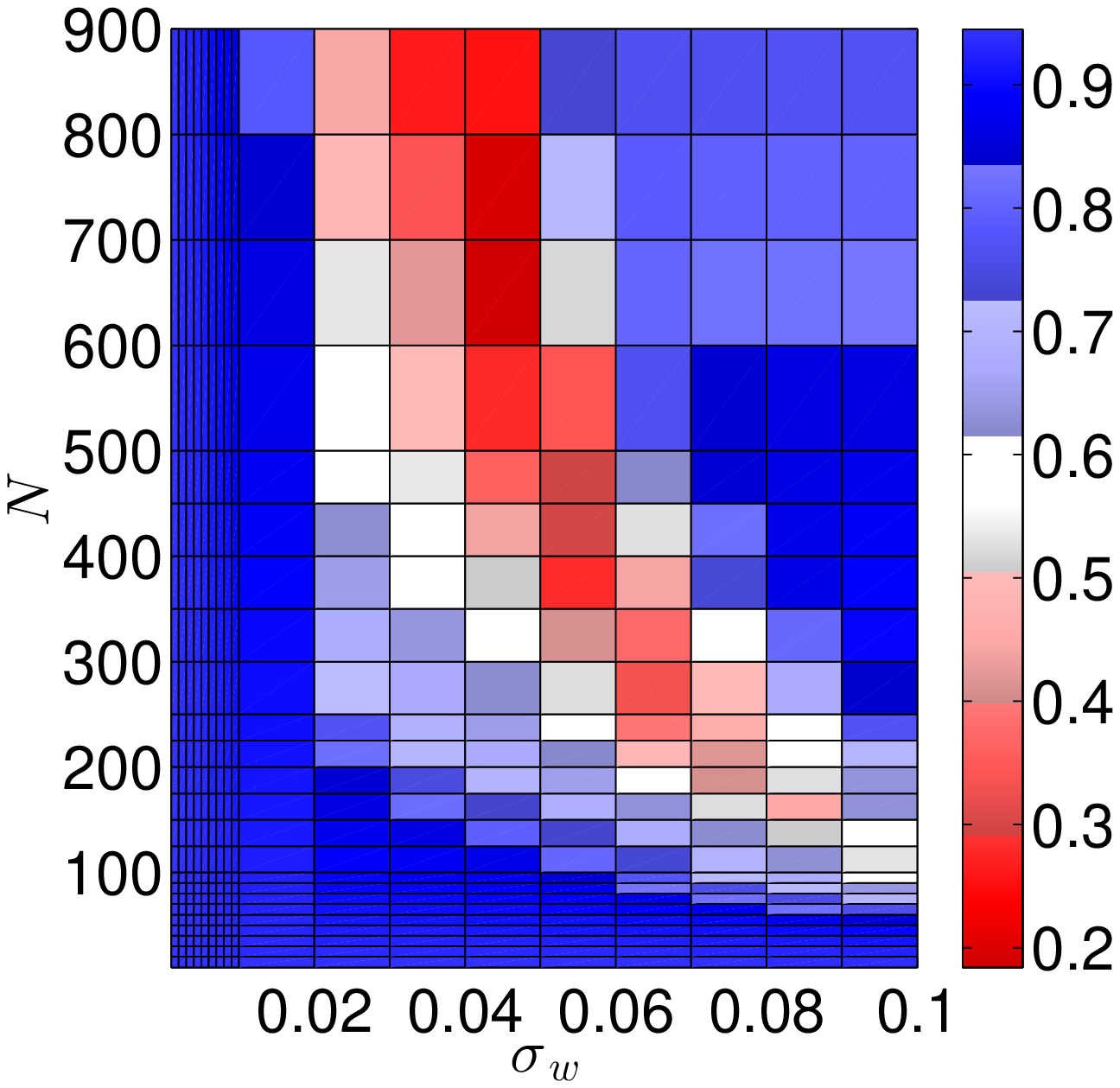}
\label{fig:surfnarma20}
}\\
\subfigure[H{\'e}non Map $\sigma_w$ scaling]{
\includegraphics[width=2.0in]{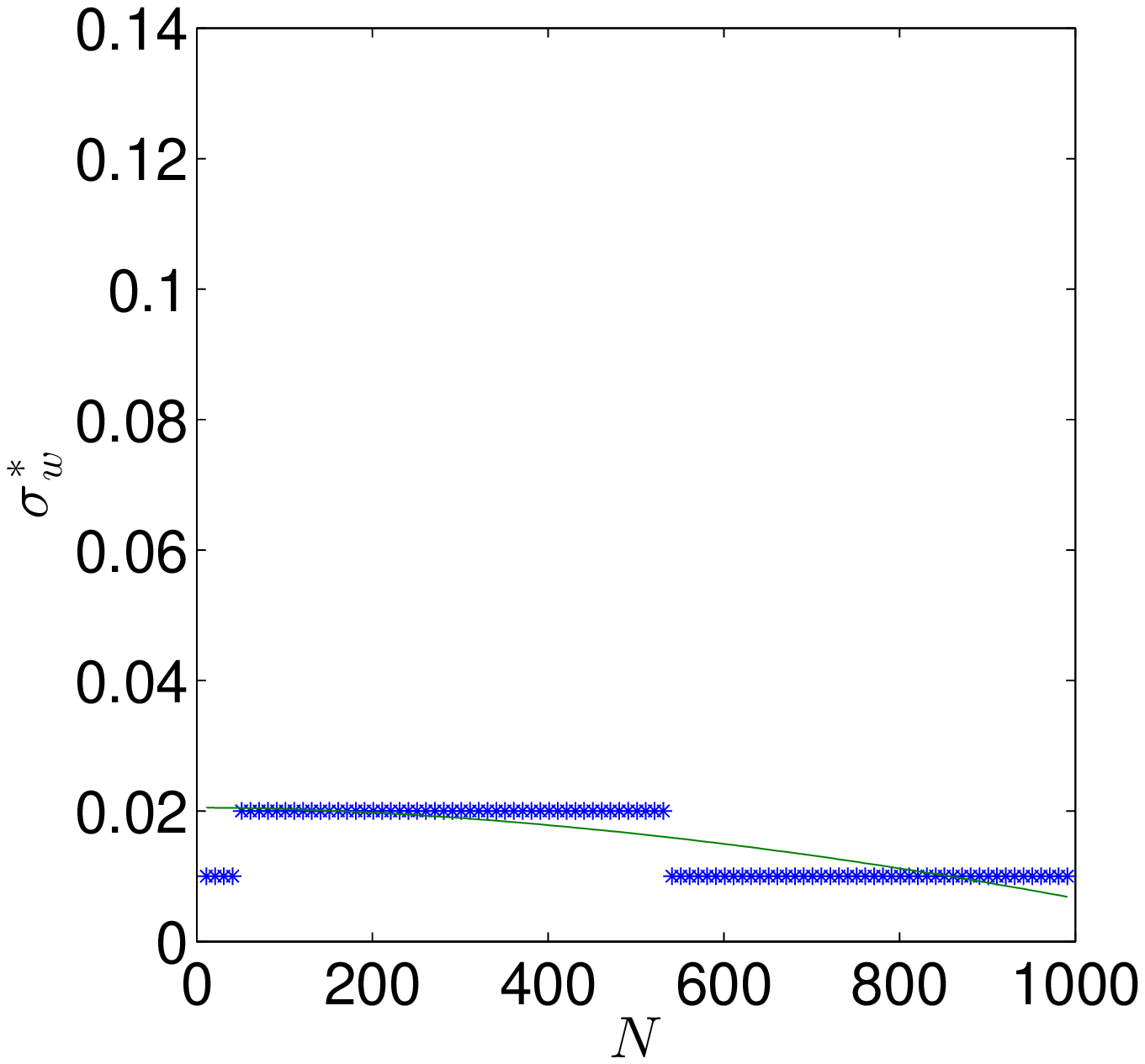}
\label{fig:opthanon}
}
\subfigure[NARMA 10  $\sigma_w$ scaling]{
\includegraphics[width=2.0in]{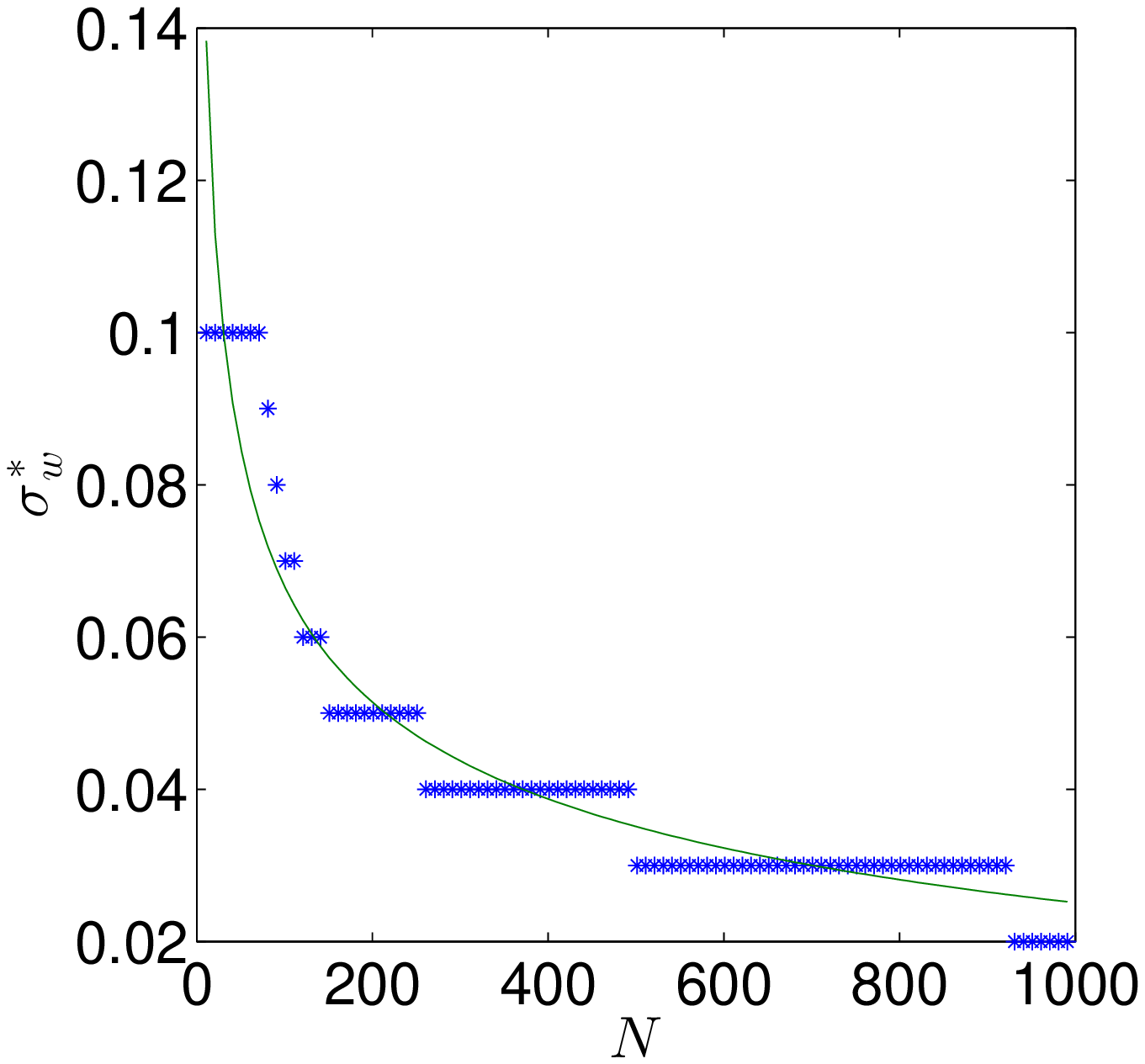}
\label{fig:optnarma10}
}
\subfigure[NARMA 20  $\sigma_w$ scaling]{
\includegraphics[width=2.0in]{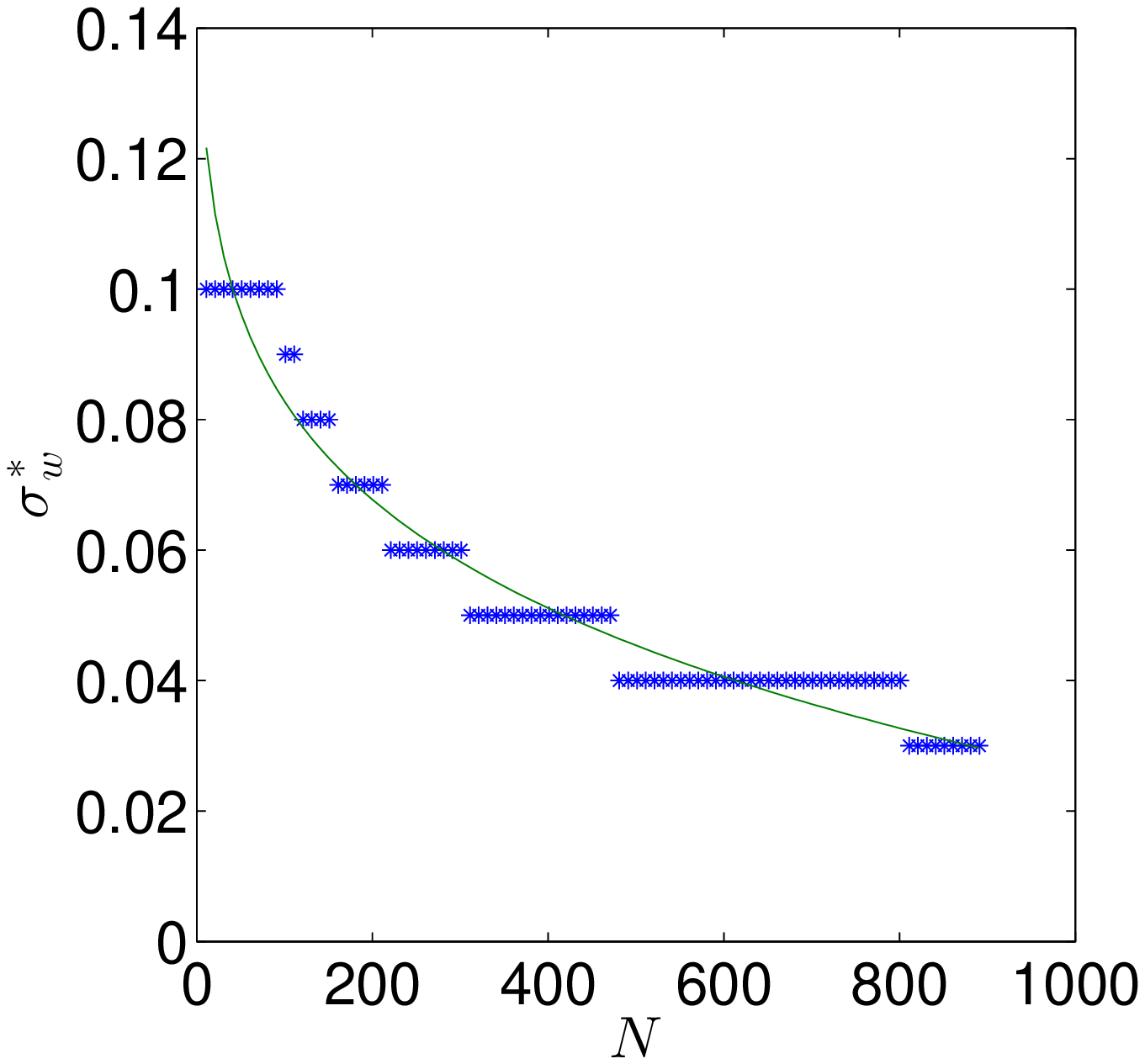}
\label{fig:optnarma20}
}
\caption{Scaling and optimization in the ESN. Figures~\ref{fig:surfhanon},~\ref{fig:surfnarma10}, and~\ref{fig:surfnarma20} show the training error surface of the ESN on the H{\'e}non Map, the NARMA 10, and the NARMA 20 tasks respectively. We create a scaling-law by finding the optimal standard deviation $\sigma_w^*(N)$ according to Equation~\ref{eq:optimize} and fitting the power-law $ax^b+c$ to it. Figures~\ref{fig:opthanon},~\ref{fig:optnarma10}, and~\ref{fig:optnarma20} show the data points (blue markers) and the fit (solid line) for the $\sigma_w^*(N)$.
}
\label{fig:optimization}
\end{figure*}

\subsection{Reservoir optimization}
\label{sec:optimization}
Depending on the ESN architecture, its performance can be sensitive to some of the 
model's parameters. These parameters are optimized using offline cross-validation 
\cite{Jaeger2007335,5629375} or online adaptation~\cite{Boedecker2009,Dasgupta:
2012fk}. This is a preliminary stage before the functional comparison. We are interested in the 
scaling of these parameters, which we  study systematically. Figures~\ref{fig:surfhanon}-\ref{fig:surfnarma20} 
shows the resulting error surface by averaging the  result of the 10 runs of each $
\sigma_w$-$N$ combination. We observe that, as the nonlinearity of the task and its 
required memory increase, ESN performance becomes more sensitive to  
changes in $\sigma_w$ and favors a more heterogeneous weight assignment (larger $
\sigma_w$). We find the optimal standard deviation $\sigma_w^*$ as a function of $N
$ for each task: 
\begin{equation}
\sigma_w^*(N) = \argmin{\sigma_w}{RNMSE(\sigma_w,N)}.
\label{eq:optimize}
\end{equation} 
We found experimentally that $\sigma_w^*(N)$ is best fitted by a power-law curve. 
The bottom row of Figure~\ref{fig:optimization} shows the result of this optimization 
and the power-law fit. The details of these fits are provided in  Appendix \ref{sec:fitdata}.
For NARMA 10 and NARMA 20 the power law is well behaved, except  
for the H{\'e}non Map which is not sensitive to  $\sigma_w$. We use $\sigma_w^*=0.02$ 
for H{\'e}non Map experiments. It is noteworthy that this power-law behavior is qualitatively 
consistent with what we expect from the theoretical result in~\cite{PhysRevE.
87.042809}, although the exact power-law coefficient is task-dependent.  

\subsection{Functional Comparison}
\label{sec:funccomp}
The division between memory and computation is not fully understood. Here, we 
attempt to compare the ESN size to the size of an equivalent device with only memory 
capacity and no computational power, and to a device with limited memory and 
theoretically arbitrary computational power. A DL of length $n$ stores the perfect 
memory of the past $n$ inputs and a NARX network with $N$ hidden units can be a 
universal approximator of any time series. We would like to compare the performance 
of a linear readout with access to the reservoir state with a linear readout with access 
to the delay line states and also to the performance of the NARX network. It is clear 
that the ESN, the DL, and the NARX network are very different, i.e., the memory 
and computational power of  ESN,  DL, and  NARX networks differ 
significantly even with identical $N$. Therefore, we perform a functional 
comparison in which we study the ESN, the DL, and the NARX network with equal 
RNMSE on the same task. For  ESN, we have two parameters:
the standard deviation of the normal distribution used to create the weight matrix
 $\sigma_w$ and the number of nodes $N$ in the reservoir. We chose the $\sigma_w$ 
that optimizes the performance of ESN for each $N$ as described in Section \ref{sec:optimization}.
\begin{figure*}[ht!]
\centering
\subfigure[H{\'e}non Map training]{
\includegraphics[width=2.0in]{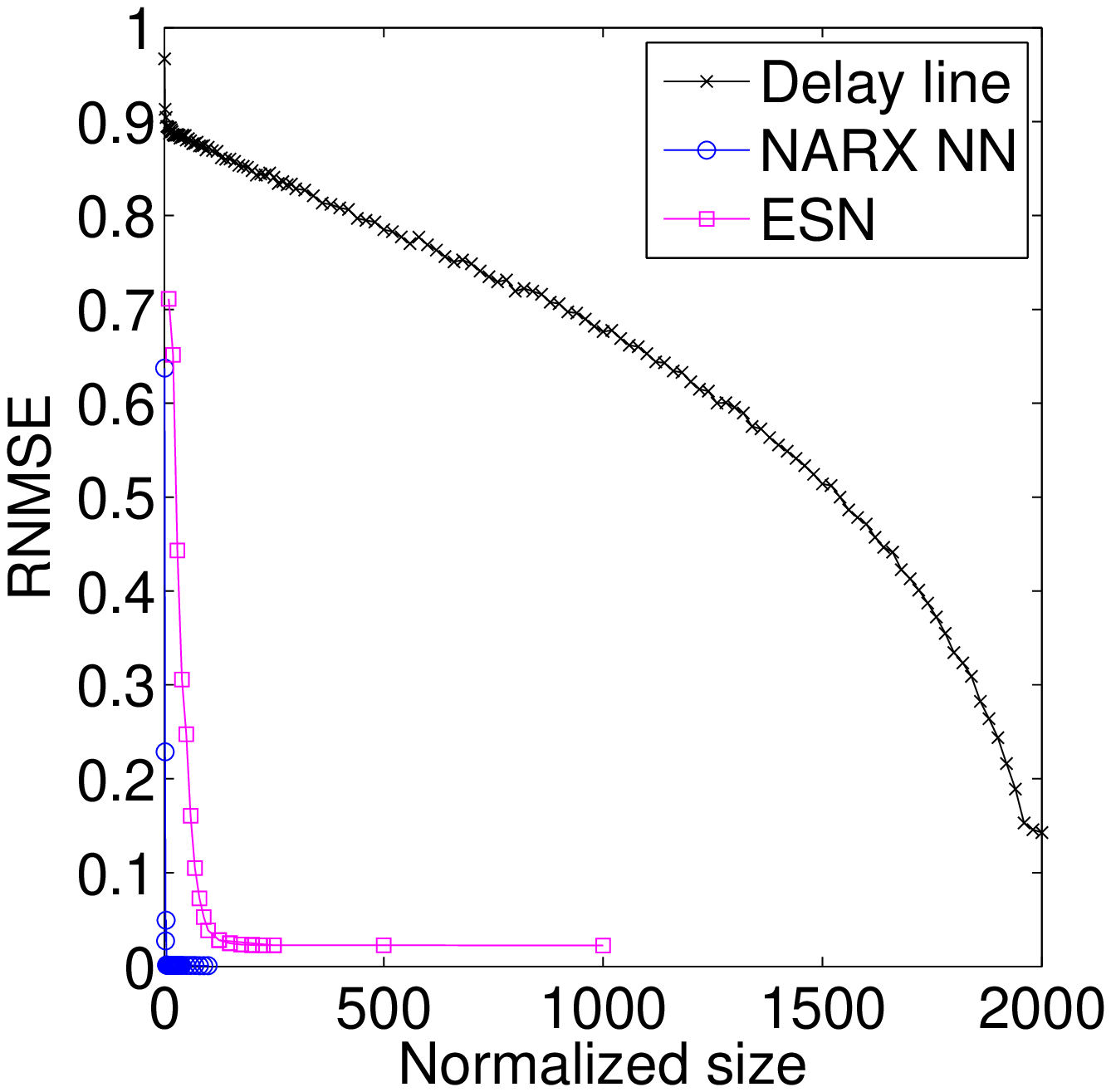}
\label{fig:henontrain}
}
\subfigure[NARMA 10 training]{
\includegraphics[width=2.0in]{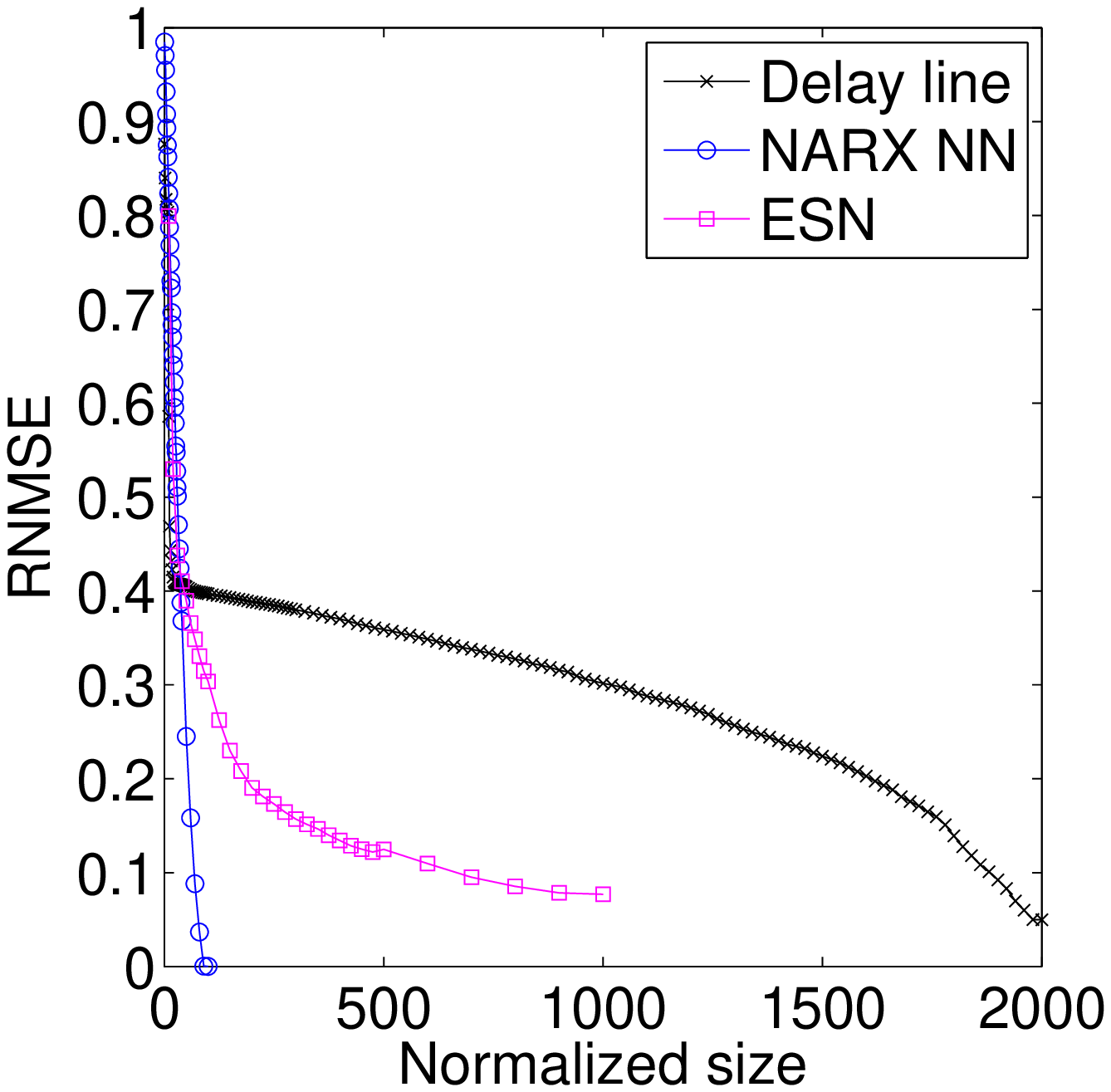}
\label{fig:narma10train}
}
\subfigure[NARMA 20 training]{
\includegraphics[width=2.0in]{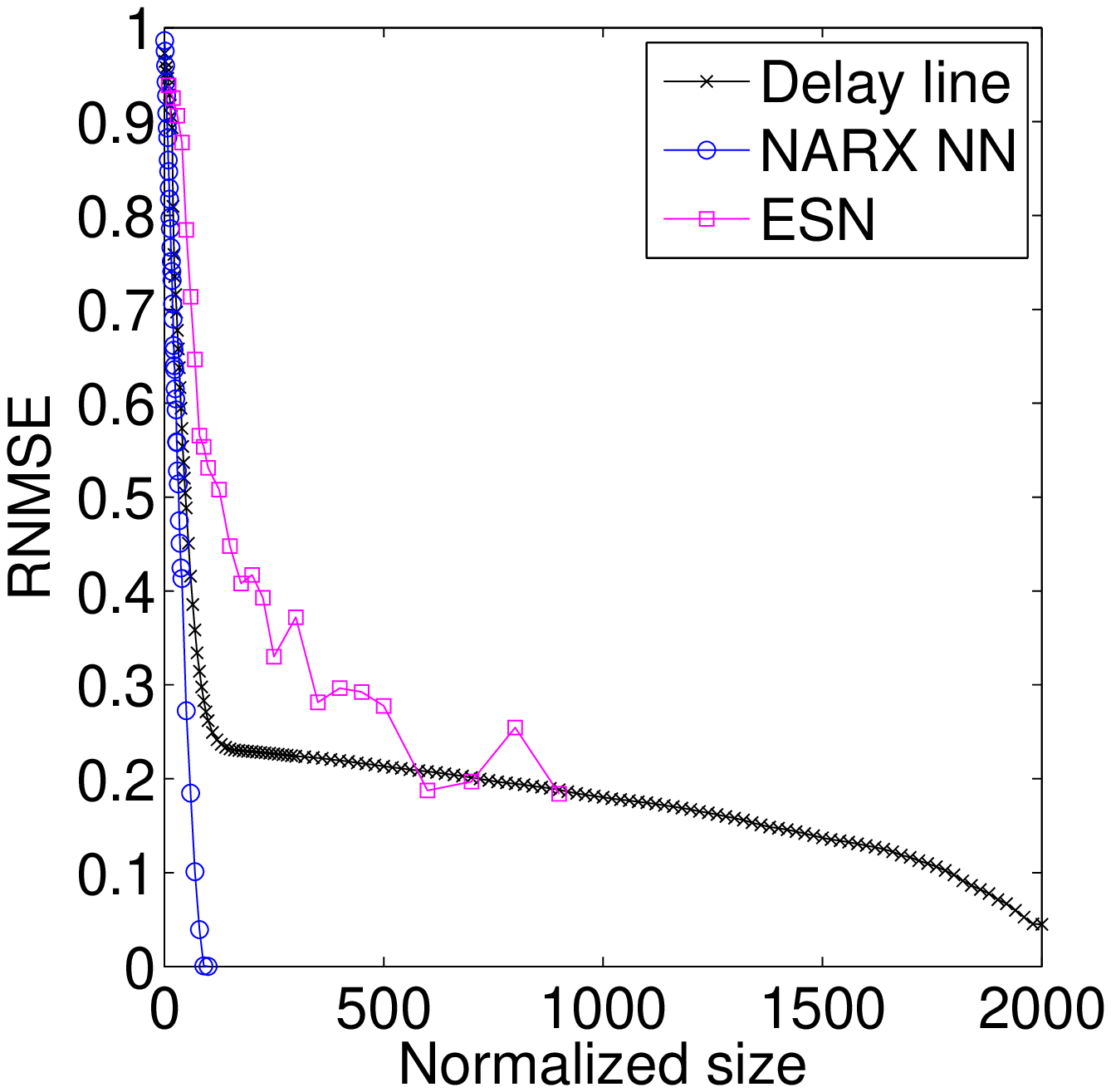}
\label{fig:narma20train}
}\\
\subfigure[H{\'e}non Map generalization]{
\includegraphics[width=2.0in]{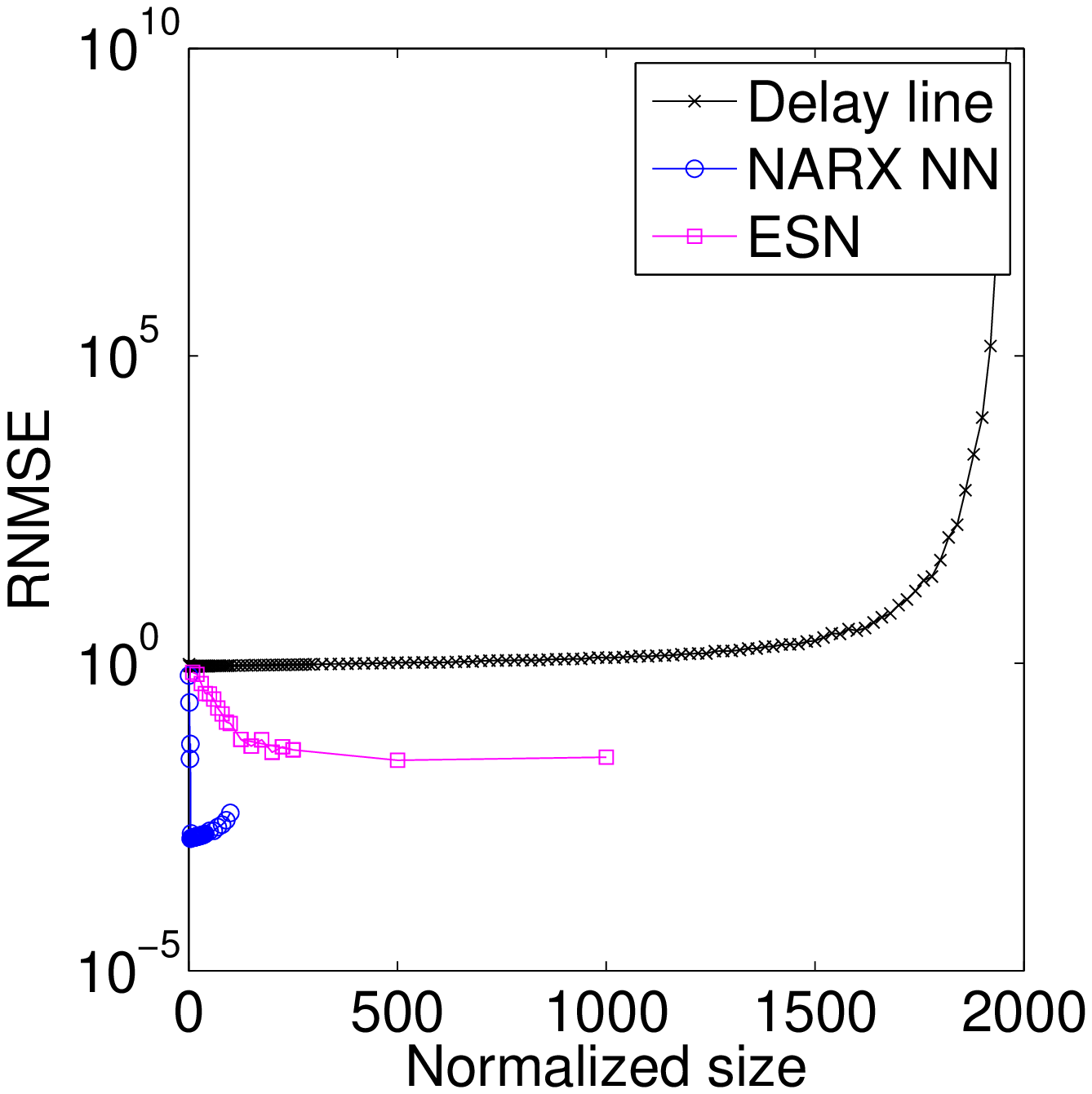}
\label{fig:henontest}
}
\subfigure[NARMA 10 generalization]{
\includegraphics[width=2.0in]{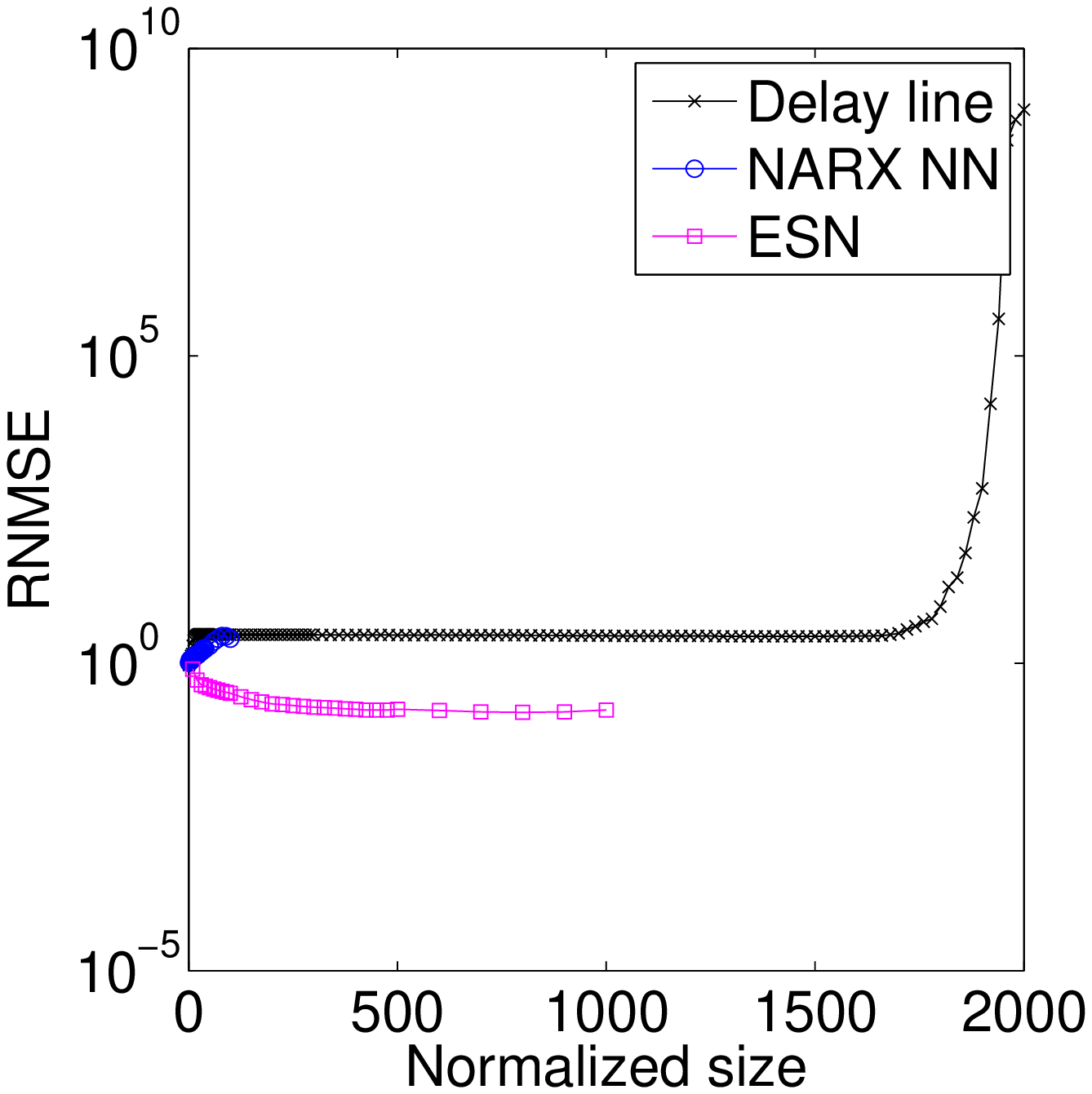}
\label{fig:narma10test}
}
\subfigure[NARMA 20 generalization]{
\includegraphics[width=2.0in]{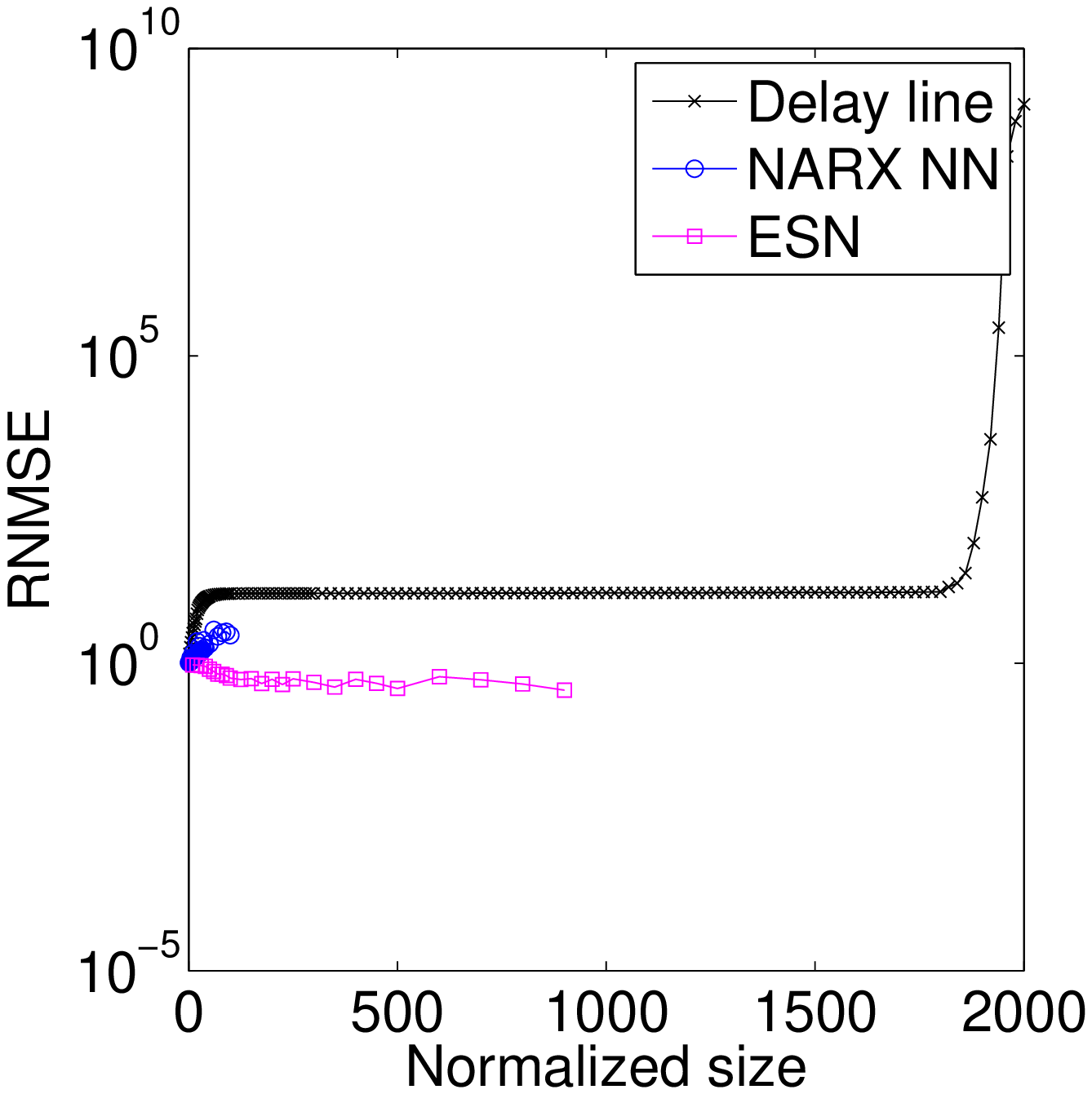}
\label{fig:narma20test}
}
\caption{Training and generalization RNMSE of the DL, the NARX network, and the ESN for three different tasks. The DL can memorize the patterns, but not generalize for any of the tasks. The NARX network can generalize for the H{\'e}non Map, but overfits for the NARMA 10 and the NARMA 20. The ESN can both memorize and generalize the temporal patterns for all the tasks.}
\label{fig:perf}
\end{figure*}
\subsection{Experimental Setup}
\label{sec:setup}
In this section we describe the parameters and data sets used for our simulations. The 
training and testing for delay line and NARX networks are done by generating 10 time series of 4,000 
time steps. We used 20 time seres of the same length for ESNs.
 We used the first 2,000 time steps of each time series to train the model 
and the second 2,000 steps to test the model. The training and testing performance 
metrics are then averaged over the 10 runs. The model specific setting are as follows:
\subsubsection{Delay line} A delay line is a fixed system with only one parameter $N$ 
that defines the number of taps. For this study we limit ourselves to $1\le N\le 2,000$.
 We have also experimented with $2,000\le N\le 3,000$, but the performance of the 
 delay line does not change for $N>2,000$. We take $N=2,000$ as the largest delay line
 for this study.
\subsubsection{NARX  neural network} Since the training in the NARX network is 
sensitive to the random initial weights, we instantiate a new network for each time series. We 
fix the number of input taps to 10 and the number of hidden layers to 1. We use the 
number of nodes in the hidden layer $N$ as the control parameter, 
with $1 \le N \le 100$.
\subsubsection{Echo state network} A single instantiation of ESN contains randomly 
assigned weights and the reservoir initial states. To average the ESN properties over 
these variations we instantiate five ESN to train and test for every time series.
The error is then averaged over the five instances and then over 20 time series. The 
variable parameters for ESN are the number of reservoir nodes $N$ and the standard 
deviation of the normal distribution used to generate the reservoir and the input 
weights. However, for each $N$ we only  use the $\sigma_w^*$ as described in 
Section~\ref{sec:optimization}. We study the ESN with the reservoir size $10 \le N \le 
1,000$.
\section{Results}
Figures~\ref{fig:henontrain}-\ref{fig:narma20train} show the 
training performance measured in RNMSE as a function of the normalized system size. In 
all three tasks, the delay line shows a sharp decrease in error as soon as the system 
acquires enough capacity to hold all the required information for the task. This is 
$N=2$ for H{\'e}non Map, $N=10$ for NARMA 10 task, and $N=20$ for NARMA 20 
task. After this point, the error decreases slowly until $N=2,000$ where $RNMSE\approx 0.14$ 
after a sharp drop. The decrease in error is due to the ``dimensionality curse": the
fixed-length teacher time series are not  representatives of the expanded state
space of the delay line. This is  
expected to result in overfitting, which is reflected in the high testing errors on  
Figures~\ref{fig:henontest}-\ref{fig:narma20test}. Another expected behavior of overfitting to training data is that if the distribution
of the data is wide, the error will be larger and for narrower distributions of data the error will be lower.
This is why the delay line has the highest error for the simplest task, the H{\'e}non Map, and the lowest
error for the most difficult task, the NARMA 20 task. 
Note that to make the testing errors for all the system readable, 
we use logarithmic  y-axes. 

The NARX network behaves differently on the H{\'e}non Map and the two NARMA 
tasks. For the H{\'e}non Map, it shows the best training and testing performance 
around $N=5$ and begins to overfits for $N>5$. For both NARMA 10 and NARMA 20, 
the training error decreases gradually  as the number of hidden nodes increases, but 
the system only memorizes the patterns and cannot generalize, which is characterized 
by increasing test RNMSE. For the NARMA 10 task, the observed training RNMSE for 
NARX network is comparable to the error of $0.17$ reported by Atiya and Parlos~
\cite{846741} for the same network size $N=40$. However, they used only 200 data 
points, which explains the slightly lower error. Atiya and Parlos focused on the 
convergence times of different algorithms and did not publish their testing error.

For the ESN, the training error on all three tasks decreases rapidly as the size of the 
reservoir increases and reaches a plateau for $N\approx 1,000$. However, unlike the 
DL and the NARX network, the testing error also decreases sharply as the reservoir 
size increases. As expected, this decrease is sharper for easier tasks.
The main different of the ESN performance is that the training error 
increases across all $N$ as the difficulty of the task increases, which 
is a sign that the readout layer is not merely memorizing the training patterns. 
 We have tabulated the error values for all three systems on all the tasks for a few system sizes, which 
can be found in the Appendix~\ref{sec:appb}. Our ESN testing results are similar to those reported 
by Rodan and Tino~\cite{5629375} for the same system size.

\begin{figure}[hb!]
\centering
\subfigure[DL vs. ESN]{
\includegraphics[width=2.5in]{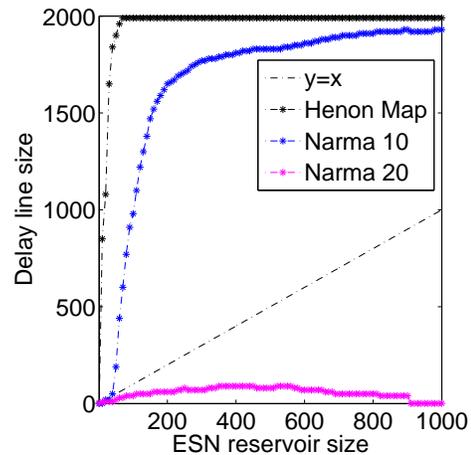}
\label{fig:dlesn}
}
\subfigure[NARX NN vs. ESN]{
\includegraphics[width=2.5in]{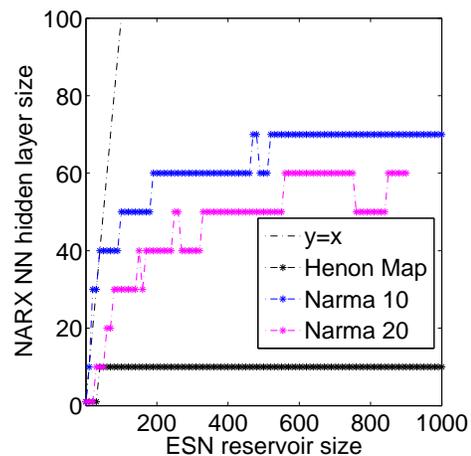}
\label{fig:narxesn}
}
\caption{Figure~\ref{fig:dlesn} shows how the complexity of the DL compares with the ESN of identical memorization performance. Except for the NARMA 20, which requires perfect memory of the 20 previous time steps, the ESN memorization capability far surpasses the DL. Figure~\ref{fig:narxesn} shows the same comparison between the NARX and the ESN. The NARX network out performs the ESN in all tasks. Due its complexity, the NARX network can memorize the patterns very well, but is not able to generalize to the new patterns (see Figure~\ref{fig:perf}).}
\label{fig:test_comp}
\end{figure}

Next, we compared the performance of the DL, the NARX network, and the ESN as 
described in Section~\ref{sec:funccomp}. Because the testing error of the systems do 
not overlap, we have to use the training error to perform a direct functional 
comparison between the three systems. This allows us to compare their 
memorization capabilities. Figure~\ref{fig:dlesn} shows the DL size as 
a function of the ESN size of identical training error for all three tasks.
For the H{\'e}non Map and the NARMA 10 tasks, the ESN achieves the same
RNMSE as the delay line with significantly fewer nodes.
   For instance for the H{\'e}non 
Map and NARMA 10, to achieve a training RNMSE of an ESN with 400 nodes, a delay
line would need $1,990$ and $1,810$ nodes respectively. For NARMA 20, the delay line
only needs $90$ nodes to achieve the same RNMSE as an ESN with $400$ nodes. 
The narrow distribution of the NARMA 20 time series allow the linear delay line 
to exploit the average case strategy to achieve a lower RNMSE. On the other hand,
the ESN readout layer learns the task itself as in contrast to just memorizing patterns. The delay line use the 
same strategy for the easier tasks as well (NARMA 10 and H{\'e}non Map), but
ESN can fit to the the training data much better than the delay line and therefore
requires much less resources to achieve the same error level.

Figure~\ref{fig:narxesn} shows the functional comparison result between NARX networks
and ESNs. A NARX network would need 10 hidden nodes to achieve the same RNMSE as a 400-node
ESN on the H{\'e}non Map task. The short time dependency and the simple form of this task
make it very easy for the network to learn the system, which results in low training and testing RNMSE.
For the NARMA 10 and the NARMA 20 tasks, the network requires 60 and 50 hidden nodes
 to be equivalent of the 400-node ESN. The strategy here is similar to the delay line where during learning
 the network tries to fit the training data on average, as best as it can. As expected this strategy will have two 
 consequences: (1) the NARMA 20 task will be easier because of its distribution; (2) the network overfits 
 the training data and cannot generalize to testing data.

\section{Discussion and Open Problems}

The reservoir in an ESN is a dynamical system in which memory and computation are 
inseparable. To understand this type of information 
processing we compared the ESN performance with a memory-only device, the DL, 
and a limited-memory but computationally powerful device, the NARX network. Our 
results illustrate that the performance of ESN is not only due to its memory capacity; ESN readout does not  create an autoregression of the input, such as in the DL or the NARX network. 
The information processing that takes place inside a reservoir is fundamentally 
different from other types of neural networks. The exact mechanism of this process 
remains to be investigated. Studying reservoir computing usually takes place by 
analyzing the systems performance for task solving with different computational and 
memory requirements. To understand the details of information processing in a 
reservoir, we have to understand the effects of the reservoir's architecture on its 
fundamental memory and computational capacity.  We also have to be able to define the classes of tasks that can be 
parametrically varied in memory requirement and nonlinearity. Our study reveals that although ESN cannot memorize patterns as well 
as a memory device or a neural network, it greatly outperforms them in generalizing to 
novel inputs. Also, increasing reservoir size in ESN  improves the performance of 
generalization, whereas in the DL or the NARX network this will result in increased 
overfitting leading to poorer generalization. One solution would be to extend
 the receiver operation characteristic (ROC) 
and receiver error characteristic (REC) curve methods to decide on the quality of 
generalization in ESN~\cite{conf/icml/BiB03,353668,Waegeman:2008:RAO:
1316089.1316327}. In the neural network community, methods based on pruning, 
regularization, cross-validation, and information criterion have been used to alleviate 
the overfitting problem~\cite{253710,Larsen:1998fk,
548336,514876,366065,Lawrence97lessonsin,329683,Moody:1994uq,Hansen:
1994kx,doi:10.1117/12.366304,doi:10.1021/ci0342472,Baum:1989p654,Amirikian:
1994p42}. Among these methods, regularization has been successfully used in 
ESNs~\cite{Wyffels20101958}. However, these methods focus on increasing the neural 
network's performance and are not suitable to quantify overfitting or to study  task 
hardness.  Another area that requires more research is the amount of training that the 
ESN requires to guarantee a certain performance, as is described in 
probably approximately correct methods~\cite{Kearns:1994:LDD:
195058.195155,Lange:1994vn,Valiant:1984p418}. To the best of our knowledge these problems 
have not been addressed in the case of  high-dimensional dynamical systems. A well 
developed theory of computation in reservoir computing needs to address all of these 
aspects. In future work, we will study some of these issues experimentally 
and based on our observations, we will attempt to develop theoretical understanding 
of computation in the reservoir computing paradigm. 

\section{Conclusion}

Reservoir computing is an approach to neural network training which has been 
successful in areas of machine learning, time series analysis, robot control, and 
sequence learning. There has been many studies aimed at understanding the working 
of RC and the factors that affect its performance. However, because of the complexity 
of the reservoir in RC, none of these studies have been completely satisfactory and 
 have often resulted in contradictory conclusions. In this paper, we compared the performance of 
three approaches to time series analysis: the delay line, the NARX network, and the 
ESN. 
These methods vary in their memory capacity and computational power. The delay 
line retains a perfect memory of the past, but does not have any computational power. 
The NARX network only has limited memory of the past, but in principle can perform 
any 
computation. Finally, the ESN does not have an explicit access to past memory, but its 
reservoir carries out computation using the implicit memory of the past represented in 
its dynamics. Using a functional comparison we showed that for simple tasks with short time
dependencies, the delay line requires more than four times as much resources that ESN requires
to achieve the same error, while the NARX network requires 40 times less resources than ESN to achieve 
equivalent error.  For tasks with long time dependencies and narrow distributions
the delay line requires less than one fourth the resources of the ESN
and the NARX network requires less then one fifth the same resources.
However, neither a delay line nor a NARX 
network can achieve the generalization power of an ESN. Many theoretical aspects of 
reservoir computing, such as the memory-computation trade-off, and the relation 
between reservoir's structure, its dynamics, and its performance remain as open 
problems. 

\section*{Acknowledgment}
The work was support by NSF grants \#1028238 and \#1028120. M.R.L. gratefully acknowledges support from the New Mexico Cancer Nanoscience and Microsystems Training Center.


%

\appendices
\section{Fitting $\sigma_w$}
\label{sec:fitdata}
Before we can fit $\sigma_w^*$, we have to interpolate the data points on the error surface with a linear fit. This allows us to use all values of $N$ and $\sigma_w$ and create a smooth fit. Table~\ref{tbl:gofsurf} shows the goodness of fit statistics for the linear fit to the error surface. Low SSE and high $R^2$ statistics on this fit shows the the surface accurately represent the data points. 
\begin{table}[h!]
\centering
\begin{tabular}{|c|c|c|}
\hline
 task & SSE & $R^2$ \\ \hline
 H{\'e}non Map &  $ 1.5486 \times 10^{-31}$ & $1$ \\ \hline
 NARMA 10 & $2.989 \times 10^{-31}$ & $1$ \\ \hline
 NARMA 20 & $7.3725 \times 10^{-31}$ & $1$\\
  \hline
\end{tabular}
\caption{Goodness of fit statistics for the interpolant fit to the error surface (Figure~\ref{fig:optimization}) as a function of reservoir $\sigma_w$ and $N$ for all three tasks.}
\label{tbl:gofsurf}
\end{table}
We then calculate the $\sigma_w^*$ corresponding to the standard deviation of the weight matrix for each $N$ that minimizes the error. We represent $\sigma_w^*$ as a function of $N$ and fit the power-law $ax^b+c$ to it. The result of the fit and the goodness of fit statistics are given in Table~\ref{tbl:gofsigma}.
\begin{table}[ht!]
\centering
\begin{tabular}{|c|c|}
\hline
 \multicolumn{2}{|c|}{H{\'e}non Map} \\ \hline 
 SSE  & $0.0011$   \\ \hline
  $R^2$ & $0.5670$ \\ \hline
   $a$ & $-5.733 \times 10^{-8}\pm 2.9430\times 10^{-7}$ \\ \hline
    $b$ & $1.795 \pm 0.7450$ \\ \hline
     $c$ &$0.02053 \pm 0.0016$ \\ \hline \hline
      \multicolumn{2}{|c|}{NARMA 10} \\ \hline 
 SSE  & $2.9686 \times 10^{-4} $  \\ \hline
  $R^2$ & $0.9926$ \\ \hline
   $a$ & $0.306 \pm 0.0095$ \\ \hline
    $b$ & $-0.2609 \pm 0.0249$ \\ \hline
     $c$ &$-0.02537 \pm 0.0079$ \\ \hline \hline
           \multicolumn{2}{|c|}{NARMA 20} \\ \hline 
 SSE  & $5.8581 \times 10^{-4}$   \\ \hline
  $R^2$ & $0.9909$ \\ \hline
   $a$ & $-0.03441 \pm 0.0143$ \\ \hline
    $b$ & $0.2156 \pm 0.0408$ \\ \hline
     $c$ & $0.1766 \pm 0.0210$ \\ \hline
\end{tabular}
\caption{Goodness of fit statistics for the power-law fit $ax^b+c$ to $\sigma_w^*$ (Figure~\ref{fig:optimization}) as a function of reservoir size $N$ for all three tasks.}
\label{tbl:gofsigma}
\end{table}

\section{The Performance Results}
\label{sec:appb}
Tables~\ref{tbl:perf}, ~\ref{tbl:perfdl}, and~\ref{tbl:perfnarx} tabulate the average testing and training errors of  optimal ESNs, delay lines, and NARX networks for the three different tasks using three different measures, i.e., RNMSE, NRMSE, and SAMP.
\begin{table*}[ht]
\scriptsize
\centering
\begin{tabular}{|c|c|c|c|c|c|c|c|}
\hline
task & \multicolumn{2}{|c|}{measure}  & $N=50, \sigma_w^*=0.02$
&  $N=100, \sigma_w^*=0.02$
&  $N=150, \sigma_w^*=0.02$
&  $N=200, \sigma_w^*=0.02$
&  $N=500, \sigma_w^*=0.02$ \\ \hline
\multirow{6}{*}{H{\'e}non Map} & \multirow{2}{*}{RNMSE} & training  & $0.2474\pm 0.0215$  & $0.0385\pm 0.0039$  & $0.0247\pm 0.0008$  & $0.0230\pm 0.0002$  & $0.0226\pm 0.0001$ \\ \cline{3-8}
 & & testing  & $0.3805\pm 0.0738$  & $0.1053\pm 0.0225$  & $0.0448\pm 0.0061$  & $0.0359\pm 0.0048$  & $0.0264\pm 0.0026$ \\ \cline{2-8}
& \multirow{2}{*}{NRMSE} & training  & $0.0696\pm 0.0060$  & $0.0109\pm 0.0011$  & $0.0069\pm 0.0002$  & $0.0065\pm 0.0000$  & $0.0063\pm 0.0000$ \\ \cline{3-8}
 & & testing  & $0.1071\pm 0.0208$  & $0.0296\pm 0.0063$  & $0.0126\pm 0.0017$  & $0.0101\pm 0.0014$  & $0.0074\pm 0.0007$ \\ \cline{2-8}
& \multirow{2}{*}{SAMP} & training  & $16.2797\pm 1.1912$  & $3.0074\pm 0.3899$  & $1.2687\pm 0.1440$  & $0.8498\pm 0.0584$  & $0.6821\pm 0.0373$ \\ \cline{3-8}
 & & testing  & $16.6023\pm 1.2046$  & $3.2470\pm 0.4199$  & $1.4182\pm 0.1577$  & $0.9797\pm 0.0672$  & $0.9441\pm 0.0527$ \\ \hline \hline
 
task & \multicolumn{2}{|c|}{measure}  & $N=50, \sigma_w^*=0.10$
&  $N=100, \sigma_w^*=0.07$
&  $N=150, \sigma_w^*=0.05$
&  $N=200, \sigma_w^*=0.05$
&  $N=500, \sigma_w^*=0.04$ \\ \hline
\multirow{6}{*}{NARMA 10} & \multirow{2}{*}{RNMSE} & training  & $0.3896\pm 0.0073$  & $0.3036\pm 0.0104$  & $0.2300\pm 0.0058$  & $0.1904\pm 0.0051$  & $0.1248\pm 0.0056$ \\ \cline{3-8}
 & & testing  & $0.4035\pm 0.0070$  & $0.3270\pm 0.0113$  & $0.2560\pm 0.0064$  & $0.2199\pm 0.0054$  & $0.1796\pm 0.0078$ \\ \cline{2-8}
& \multirow{2}{*}{NRMSE} & training  & $0.0635\pm 0.0012$  & $0.0495\pm 0.0017$  & $0.0374\pm 0.0010$  & $0.0309\pm 0.0008$  & $0.0202\pm 0.0009$ \\ \cline{3-8}
 & & testing  & $0.0653\pm 0.0011$  & $0.0529\pm 0.0018$  & $0.0414\pm 0.0011$  & $0.0355\pm 0.0009$  & $0.0291\pm 0.0013$ \\ \cline{2-8}
& \multirow{2}{*}{SAMP} & training  & $4.6137\pm 0.0819$  & $3.5742\pm 0.1343$  & $2.6224\pm 0.0742$  & $2.1247\pm 0.0642$  & $1.4310\pm 0.0654$ \\ \cline{3-8}
 & & testing  & $4.7961\pm 0.0862$  & $3.8297\pm 0.1467$  & $2.8827\pm 0.0881$  & $2.4024\pm 0.0719$  & $1.9710\pm 0.0900$ \\ \hline \hline

task & \multicolumn{2}{|c|}{measure}  & $N=50, \sigma_w^*=0.10$
&  $N=100, \sigma_w^*=0.09$
&  $N=150, \sigma_w^*=0.08$
&  $N=200, \sigma_w^*=0.07$
&  $N=500, \sigma_w^*=0.04$ \\ \hline
\multirow{6}{*}{NARMA 20} & \multirow{2}{*}{RNMSE} & training  & $0.7846\pm 0.0246$  & $0.5313\pm 0.0510$  & $0.4478\pm 0.0650$  & $0.4171\pm 0.0621$  & $0.2776\pm 0.0183$ \\ \cline{3-8}
 & & testing  & $0.8003\pm 0.0257$  & $0.5746\pm 0.0499$  & $8.6177\pm 24.2443$  & $5.7683\pm 13.0557$  & $0.3873\pm 0.0222$ \\ \cline{2-8}
& \multirow{2}{*}{NRMSE} & training  & $0.0985\pm 0.0031$  & $0.0667\pm 0.0065$  & $0.0562\pm 0.0082$  & $0.0524\pm 0.0078$  & $0.0349\pm 0.0023$ \\ \cline{3-8}
 & & testing  & $0.1012\pm 0.0032$  & $0.0728\pm 0.0062$  & $1.0662\pm 2.9647$  & $0.7260\pm 1.6434$  & $0.0491\pm 0.0028$ \\ \cline{2-8}
& \multirow{2}{*}{SAMP} & training  & $2.9507\pm 0.0776$  & $2.1860\pm 0.1922$  & $1.9531\pm 0.2518$  & $1.8459\pm 0.2461$  & $1.3097\pm 0.0807$ \\ \cline{3-8}
 & & testing  & $3.0570\pm 0.0830$  & $2.3736\pm 0.1874$  & $6.0811\pm 7.7539$  & $5.3690\pm 7.2349$  & $1.8134\pm 0.0965$ \\ \hline

\end{tabular}
\caption{Training and testing errors for the optimal ESN on three different tasks measured using three different error metrics RNMSE, NRMSE, and SAMP. The optimal $\sigma_w^*$ is measured using the RNMSE on the training data. Each data point is averaged over 100 experiments.}
\label{tbl:perf}
\end{table*}
\begin{table*}[ht!]
\scriptsize
\centering
\begin{tabular}{|c|c|c|c|c|c|c|c|}
\hline
task & \multicolumn{2}{|c|}{measure}  & $N=100$
&  $N=200$
&  $N=500$
&  $N=1000$
&  $N=2000$ \\ \hline
\multirow{6}{*}{H{\'e}non Map}  & \multirow{2}{*}{RNMSE} & training  & $0.8726\pm 0.0044$  & $0.8478\pm 0.0038$  & $0.7848\pm 0.0049$  & $0.6764\pm 0.0060$  & $0.1426\pm 0.0066$ \\ \cline{3-8}
 & & testing  & $0.9164\pm 0.0051$  & $0.9378\pm 0.0037$  & $1.0243\pm 0.0120$  & $1.2294\pm 0.0183$  & $13305472391.5757\pm 9380091076.0147$ \\ \cline{2-8}
& \multirow{2}{*}{NRMSE} & training  & $0.2452\pm 0.0011$  & $0.2382\pm 0.0009$  & $0.2208\pm 0.0018$  & $0.1901\pm 0.0011$  & $0.0402\pm 0.0018$ \\ \cline{3-8}
 & & testing  & $0.2581\pm 0.0017$  & $0.2636\pm 0.0017$  & $0.2886\pm 0.0037$  & $0.3463\pm 0.0046$  & $3748747654.4792\pm 2646155454.1636$ \\ \cline{2-8}
& \multirow{2}{*}{SAMP} & training  & $56.0839\pm 0.3592$  & $54.1909\pm 0.3698$  & $50.1493\pm 0.3639$  & $42.7874\pm 0.4785$  & $2.5302\pm 0.1497$ \\ \cline{3-8}
 & & testing  & $57.8635\pm 0.3953$  & $58.3511\pm 0.2825$  & $59.4723\pm 0.5885$  & $60.9760\pm 0.5650$  & $95.2081\pm 0.4781$ \\ \hline \hline

task & \multicolumn{2}{|c|}{measure}  & $N=100$
&  $N=200$
&  $N=500$
&  $N=1000$
&  $N=2000$ \\ \hline
\multirow{6}{*}{NARMA 10}  & \multirow{2}{*}{RNMSE} & training  & $0.3972\pm 0.0188$  & $0.3884\pm 0.0193$  & $0.3590\pm 0.0224$  & $0.3017\pm 0.0230$  & $0.0497\pm 0.0072$ \\ \cline{3-8}
 & & testing  & $2.9150\pm 0.1827$  & $2.9088\pm 0.1852$  & $2.8797\pm 0.1919$  & $2.8063\pm 0.2152$  & $1007103272.9273\pm 482045875.2627$ \\ \cline{2-8}
 & \multirow{2}{*}{NRMSE} & training  & $0.0650\pm 0.0021$  & $0.0635\pm 0.0020$  & $0.0586\pm 0.0021$  & $0.0493\pm 0.0025$  & $0.0082\pm 0.0013$ \\ \cline{3-8}
 & & testing  & $0.4642\pm 0.0360$  & $0.4633\pm 0.0363$  & $0.4586\pm 0.0370$  & $0.4468\pm 0.0401$  & $164091125.1086\pm 83497086.3948$ \\ \cline{2-8}
 & \multirow{2}{*}{SAMP} & training  & $4.6786\pm 0.1660$  & $4.5824\pm 0.1747$  & $4.2474\pm 0.1970$  & $3.5989\pm 0.2575$  & $0.1873\pm 0.0222$ \\ \cline{3-8}
 & & testing  & $27.5133\pm 1.0130$  & $27.4396\pm 1.0251$  & $27.2029\pm 1.0782$  & $26.5114\pm 1.2347$  & $100.0000\pm 0.0000$ \\ \hline \hline

task & \multicolumn{2}{|c|}{measure}  & $N=100$
&  $N=200$
&  $N=500$
&  $N=1000$
&  $N=2000$ \\ \hline
\multirow{6}{*}{NARMA 20}  & \multirow{2}{*}{RNMSE} & training  & $0.2620\pm 0.0093$  & $0.2289\pm 0.0068$  & $0.2132\pm 0.0066$  & $0.1804\pm 0.0091$  & $0.0450\pm 0.0061$ \\ \cline{3-8}
 & & testing  & $13.5807\pm 0.7275$  & $13.6650\pm 0.7321$  & $13.7338\pm 0.7514$  & $13.9180\pm 0.7712$  & $1234560367.6547\pm 811645432.8192$ \\ \cline{2-8}
 & \multirow{2}{*}{NRMSE} & training  & $0.0333\pm 0.0020$  & $0.0290\pm 0.0014$  & $0.0271\pm 0.0014$  & $0.0229\pm 0.0017$  & $0.0057\pm 0.0007$ \\ \cline{3-8}
 & & testing  & $1.7237\pm 0.0382$  & $1.7346\pm 0.0402$  & $1.7433\pm 0.0429$  & $1.7667\pm 0.0457$  & $160391618.4866\pm 109709916.7516$ \\ \cline{2-8}
 & \multirow{2}{*}{SAMP} & training  & $1.3211\pm 0.0855$  & $1.1533\pm 0.0600$  & $1.0761\pm 0.0615$  & $0.8980\pm 0.0743$  & $0.1113\pm 0.0132$ \\ \cline{3-8}
 & & testing  & $43.8473\pm 0.5246$  & $44.0038\pm 0.5579$  & $44.1198\pm 0.5911$  & $44.4313\pm 0.6170$  & $100.0000\pm 0.0000$ \\ \hline

\end{tabular}
\caption{Training and testing errors for the delay line reservoir on three different tasks measured using three different error metrics RNMSE, NRMSE, and SAMP. Each data point is averaged over 10 experiments.}
\label{tbl:perfdl}
\end{table*}
\begin{table*}[ht!]
\scriptsize
\centering
\begin{tabular}{|c|c|c|c|c|c|c|c|}
\hline
task & \multicolumn{2}{|c|}{measure}  & $N=5$
&  $N=10$
&  $N=30$
&  $N=50$
&  $N=100$ \\ \hline
\multirow{6}{*}{H{\'e}non Map} & \multirow{2}{*}{RNMSE} & training  & $0.0014\pm 0.0000$  & $0.0013\pm 0.0000$  & $0.0012\pm 0.0000$  & $0.0011\pm 0.0000$  & $0.0009\pm 0.0000$ \\ \cline{3-8}
 & & testing  & $0.0014\pm 0.0000$  & $0.0015\pm 0.0000$  & $0.0016\pm 0.0000$  & $0.0019\pm 0.0001$  & $0.0037\pm 0.0008$ \\ \cline{2-8}
& \multirow{2}{*}{NRMSE} & training  & $0.0004\pm 0.0000$  & $0.0004\pm 0.0000$  & $0.0003\pm 0.0000$  & $0.0003\pm 0.0000$  & $0.0003\pm 0.0000$ \\ \cline{3-8}
 & & testing  & $0.0004\pm 0.0000$  & $0.0004\pm 0.0000$  & $0.0005\pm 0.0000$  & $0.0005\pm 0.0000$  & $0.0010\pm 0.0002$ \\ \cline{2-8}
& \multirow{2}{*}{SAMP} & training  & $0.1900\pm 0.0304$  & $0.1692\pm 0.0292$  & $0.1674\pm 0.0293$  & $0.1455\pm 0.0140$  & $0.1098\pm 0.0142$ \\ \cline{3-8}
 & & testing  & $0.2091\pm 0.0293$  & $0.1914\pm 0.0218$  & $0.2110\pm 0.0294$  & $0.2323\pm 0.0354$  & $0.3263\pm 0.0335$ \\ \hline \hline
 
 \multirow{6}{*}{NARMA 10} & \multirow{2}{*}{RNMSE} & training  & $0.9081\pm 0.0072$  & $0.8235\pm 0.0091$  & $0.5010\pm 0.0094$  & $0.2450\pm 0.0100$  & $0.0000\pm 0.0000$ \\ \cline{3-8}
 & & testing  & $1.0894\pm 0.0074$  & $1.1627\pm 0.0121$  & $1.5363\pm 0.0233$  & $1.9322\pm 0.0330$  & $2.5015\pm 0.0755$ \\ \cline{2-8}
& \multirow{2}{*}{NRMSE} & training  & $0.2617\pm 0.0018$  & $0.2373\pm 0.0028$  & $0.1444\pm 0.0026$  & $0.0706\pm 0.0028$  & $0.0000\pm 0.0000$ \\ \cline{3-8}
 & & testing  & $0.3156\pm 0.0030$  & $0.3368\pm 0.0048$  & $0.4451\pm 0.0080$  & $0.5598\pm 0.0110$  & $0.7246\pm 0.0227$ \\ \cline{2-8}
& \multirow{2}{*}{SAMP} & training  & $26.2977\pm 0.3168$  & $24.2035\pm 0.3464$  & $17.0853\pm 0.2634$  & $10.3269\pm 0.4612$  & $0.0000\pm 0.0000$ \\ \cline{3-8}
 & & testing  & $30.4646\pm 0.4856$  & $31.9513\pm 0.5791$  & $40.4312\pm 0.8923$  & $47.1503\pm 0.6935$  & $52.7679\pm 1.0017$ \\ \hline \hline
 
 \multirow{6}{*}{NARMA 20} & \multirow{2}{*}{RNMSE} & training  & $0.9276\pm 0.0046$  & $0.8468\pm 0.0063$  & $0.5279\pm 0.0100$  & $0.2725\pm 0.0084$  & $0.0000\pm 0.0000$ \\ \cline{3-8}
 & & testing  & $1.2201\pm 0.2734$  & $1.4458\pm 0.5060$  & $1.5594\pm 0.0238$  & $2.0689\pm 0.0580$  & $2.8666\pm 0.0921$ \\ \cline{2-8}
& \multirow{2}{*}{NRMSE} & training  & $0.2673\pm 0.0015$  & $0.2441\pm 0.0022$  & $0.1521\pm 0.0030$  & $0.0785\pm 0.0026$  & $0.0000\pm 0.0000$ \\ \cline{3-8}
 & & testing  & $0.3538\pm 0.0806$  & $0.4189\pm 0.1466$  & $0.4517\pm 0.0077$  & $0.5994\pm 0.0186$  & $0.8307\pm 0.0310$ \\ \cline{2-8}
& \multirow{2}{*}{SAMP} & training  & $26.6541\pm 0.2840$  & $24.7928\pm 0.2442$  & $17.6939\pm 0.2953$  & $11.0069\pm 0.4056$  & $0.0000\pm 0.0000$ \\ \cline{3-8}
 & & testing  & $30.4213\pm 0.6264$  & $31.9725\pm 0.7206$  & $40.2941\pm 0.7937$  & $47.8871\pm 0.8487$  & $54.5926\pm 1.1815$ \\ \hline
\end{tabular}
\caption{Training and testing errors for the NARX network on three different tasks measured using three different error metrics RNMSE, NRMSE, and SAMP. Each data point is averaged over 10 experiments.}
\label{tbl:perfnarx}
\end{table*}

\bibliographystyle{IEEEtran}
\bibliography{rc}

\end{document}